\def\year{2019}\relax
\documentclass[letterpaper]{article} 
\usepackage{aaai19}  
\usepackage{helvet}  
\usepackage{courier}  
\usepackage{url}  

\usepackage{times}
\usepackage{epsfig}
\usepackage{graphicx}
\usepackage{subfig}
\usepackage{amsmath,amssymb,amsfonts}
\usepackage{nicefrac}       
\usepackage{xcolor}
\usepackage{multirow}
\usepackage{array}
\usepackage{algorithm, algorithmicx, algpseudocode}
\usepackage{hhline}

\graphicspath{{data/}}

\newtheorem{theorem}{Theorem}

\newtheorem{corollary}{Corollary}
\newtheorem{lemma}{Lemma}
\newtheorem{definition}{Definition}

\begin{document}

%
\title{Composite Binary Decomposition Networks}
\author{You Qiaoben$^1$,
Zheng Wang$^2$,
Jianguo Li$^3$,
Yinpeng Dong$^1$,
Yu-Gang Jiang$^2$,
Jun Zhu$^1$\\
\textsuperscript{1}{Dept. of Comp. Sci. \& Tech., State Key Lab for Intell. Tech. \& Sys., Institute for AI, Tsinghua University}\\
\textsuperscript{2}{School of Computer Science, Fudan University}\\
\textsuperscript{3}{Intel Labs China}\\
qby\_222@126.com,
\{zhengwang17,ygj\}@fudan.edu.cn,
jianguo.li@intel.com,
\{dyp17@mails., dcszj@\}tsinghua.edu.cn}

\maketitle

\begin{abstract}
Binary neural networks have great resource and computing efficiency,
while suffer from long training procedure and non-negligible accuracy drops, when comparing to the full-precision counterparts.
In this paper, we propose the composite binary decomposition networks (CBDNet), which first compose real-valued tensor of each layer with a limited number of binary tensors,
and then decompose some conditioned binary tensors into two low-rank binary tensors,
so that the number of parameters and operations are greatly reduced comparing to the original ones.
%
Experiments demonstrate the effectiveness of the proposed method, as CBDNet can approximate image classification network ResNet-18 using 5.25 bits, VGG-16 using 5.47 bits,
DenseNet-121 using 5.72 bits, object detection networks SSD300 using 4.38 bits, and semantic segmentation networks SegNet using 5.18 bits, all with minor accuracy drops. \footnote{This work was done when You Qiaoben and Zheng Wang were interns at Intel Labs. Jianguo Li is the corresponding author.}
\end{abstract}

\section{Introduction}
With the remarkable improvements of Convolutional Neural Networks (CNNs), varied excellent performance has been achieved in a wide range of pattern recognition tasks, such as image classification~\cite{krizhevsky2012imagenet,szegedy2015going,he2015deep,densenet}, object detection~\cite{girshick2014rich,ren2015faster,shen2017dsod} and semantic segmentation~\cite{long2015fully,badrinarayanan2017segnet}, etc.
A well-performed CNN based systems usually need considerable storage and computation power to store and calculate millions of parameters in tens or even hundreds of CNN layers.
Therefore,the deployment of CNNs to some resource limited scenarios is hindered, especially low-power embedded devices in the emerging Internet-of-Things (IoT) domain.

Many efforts have been devoted to optimizing the inference resource requirement of CNNs, which can be roughly divided into three categories according to the life cycle of deep models.
\textit{First, design-time network optimization} considers designing efficient network structures from scratch in a handcraft way such as MobileNet~\cite{howard2017mobilenets}, interlacing/shuffle networks~\cite{zhang2017igc,zhang2017shufflenet},
 or even automatic search way such as NASNet~\cite{zoph2016neural}, PNASNet~\cite{liu2017pnas}.
\textit{Second, training-time network optimization} tries to simplify the pre-defined network structures on neural connections \cite{han-learning,deep-compression}, filter structures \cite{Wen2016Learning,Li2016Pruning,liu2017learning}, and even weight precisions \cite{hashnet,binarynet,rastegari2016xnor}
through regularized retraining or fine-tuning or even knowledge distilling \cite{hinton2015distilling}.
\textit{Third, deploy-time network optimization} tries to replace heavy/redundant components/structures in pre-trained CNN models with efficient/lightweight
ones in a training-free way. Typical works include low-rank decomposition \cite{Denton2014Exploiting}, spatial decomposition \cite{Jaderberg2014Speeding},
channel decomposition \cite{Zhang2016Accelerating} and network decoupling \cite{guo2018nd}.

To produce desired outputs, it is obvious that the first two categories of methods require a time-consuming training procedure with full training-set available, while methods of the third category may not require training-set, or in some cases require a small dataset (e.g., 5000 images) to calibrate some parameters.
The optimization process can be typically done within dozens of minutes.
Therefore, in case that the customers can't provide training data due to privacy or confidential issues, it is of great value when software/hardware vendors help their customers optimize CNN based solutions.  It also opens the possibility for on-device learning to compression, and online learning with new ingress data.
In consequence, there is a strong demand for modern deep learning frameworks or hardware (GPU/ASIC/FPGA) vendors to provide deploy-time model optimizing tools.

\begin{figure}[tp!]
\centering
\small
\includegraphics[width=0.99\linewidth]{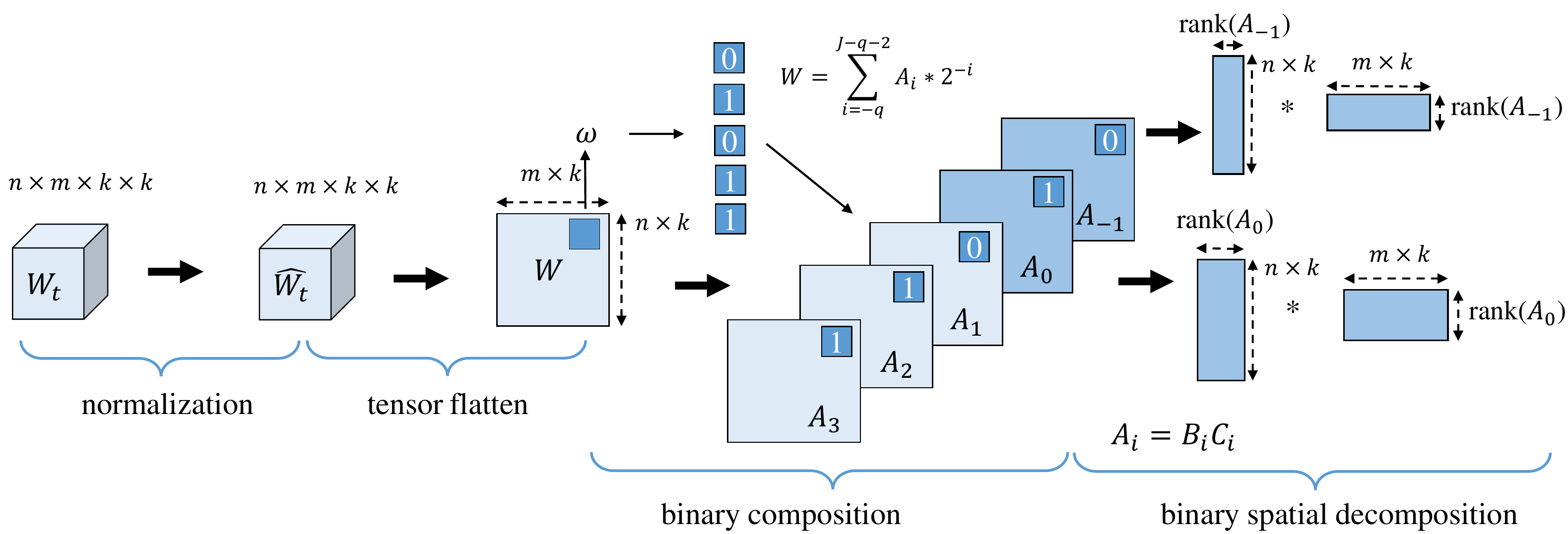}
\caption{Overall framework illustration of CBDNet.}
\label{fig:cbd}
\end{figure}
However, current deploy-time optimization methods can only provide very limited optimization (2$\sim$4$\times$ in compression/speedup) over original models.
Meanwhile, binary neural networks~\cite{courbariaux2015binaryconnect,binarynet}, which aim for training CNNs with binary weights or even binary activations, attract much more attention due to their high compression rate and computing efficiency.
However, binary networks generally suffer much from a long training procedure and non-negligible accuracy drops, when comparing to the full-precision (FP32) counterparts.
Many efforts have been spent to alleviate this problem in training-time optimization \cite{rastegari2016xnor,zhou2016dorefa}.
This paper considers the problem from a different perspective via raising the question: \textit{is it possible to directly transfer full-precision networks into binary networks at deploy-time in a training-free way?}
We study this problem, and give a positive answer by proposing a solution named composite binary decomposition networks (CBDNet).
Figure \ref{fig:cbd} illustrates the overall framework of the proposed method.
The main contributions of this paper are summarized as below:
\begin{itemize}
    \item We show that full-precision CNN models can be directly transferred into highly parameter and computing efficient multi-bits binary network models in a training-free way by the proposed CBDNet.
    \item We propose an algorithm to first expand full-precision tensors of each conv-layer with a limited number of binary tensors,
    and then decompose some conditioned binary tensors into two low-rank binary tensors.
    To our best knowledge, we are the first to study the network sparsity and the low-rank decomposition in the binary space.
    \item We demonstrate the effectiveness of CBDNet on different classification networks including VGGNet, ResNet, DenseNet as well as detection network SSD300 and semantic segmentation network SegNet. This verifies that CBDNet is widely applicable.
\end{itemize}


\section{Related Work}
\subsection{Binary Neural Networks}
Binary neural networks~\cite{courbariaux2015binaryconnect,binarynet,rastegari2016xnor} with high compression rate and great computing efficiency, have progressively attracted attentions owing to their great inference performance.

Particularly, BinaryConnect (BNN)~\cite{courbariaux2015binaryconnect} binarizes weights to $+1$ and $-1$ and  substitutes multiplications with additions and subtractions to speed up the computation.
As well as binarizing weight values plus one scaling factor for each filter channel,
Binary weighted networks (BWN) \cite{rastegari2016xnor} extends it to XNOR-Net with both weights and activations binarized.
DoReFaNet \cite{zhou2016dorefa} binarizes not merely weights and activations, but also gradients for the purpose of fast training.
However, binary networks are facing the challenge that accuracy may drops non-negligibly, especially for very deep models (e.g., ResNet).
In spite of the fact that \cite{hou2016loss} directly consider the loss to mitigate possible accuracy drops to mitigate during binarization, which gain more accurate results than BWN and XNOR-Net, it still has gap to the full-precision counterparts. 
A novel training procedure named stochastic quantization \cite{dong2017sq} was introduced to narrow down such gaps. 
All these works belongs to the training-time optimization category in summary.

\subsection{Deploy-time Network Optimization}
Deploy-time network optimization tries to replace some heavy CNN structures in pre-trained CNN models with efficient ones in a training-free way.
Low-rank decomposition \cite{Denton2014Exploiting} exploits low-rank nature within CNN layers, and shows that fully-connected (FC) layers can be efficiently compressed and accelerated with low-rank approximations, while conv-layers can not.
Spatial decomposition \cite{Jaderberg2014Speeding} factorizes the $k_h$$\times$ $k_w$  convolutional filters into a linear combination of a horizontal filter 1$\times$$k_w$ and a vertical filter $k_h$$\times$ 1.
Channel decomposition \cite{Zhang2016Accelerating} decomposes one conv-layer into two layers, while the first layer has the same filter-size but
with less channels, and the second layer uses a 1$\times$1 convolution to mix output of the first one.
Network decoupling \cite{guo2018nd} decomposes the regular convolution into the successive combination of depthwise convolution and pointwise convolution. 

Due to its simplicity, deploy-time optimization has many potential applications for software/hardware vendors as aforementioned.
However, it suffers from relatively limited optimization gains (2$\sim$4$\times$ in compression/speedup) over original full-precision models.

\begin{figure*}[tph]
\begin{minipage}{.495\linewidth}
\centering
\small
\subfloat[]{
\includegraphics[width=.5\linewidth]{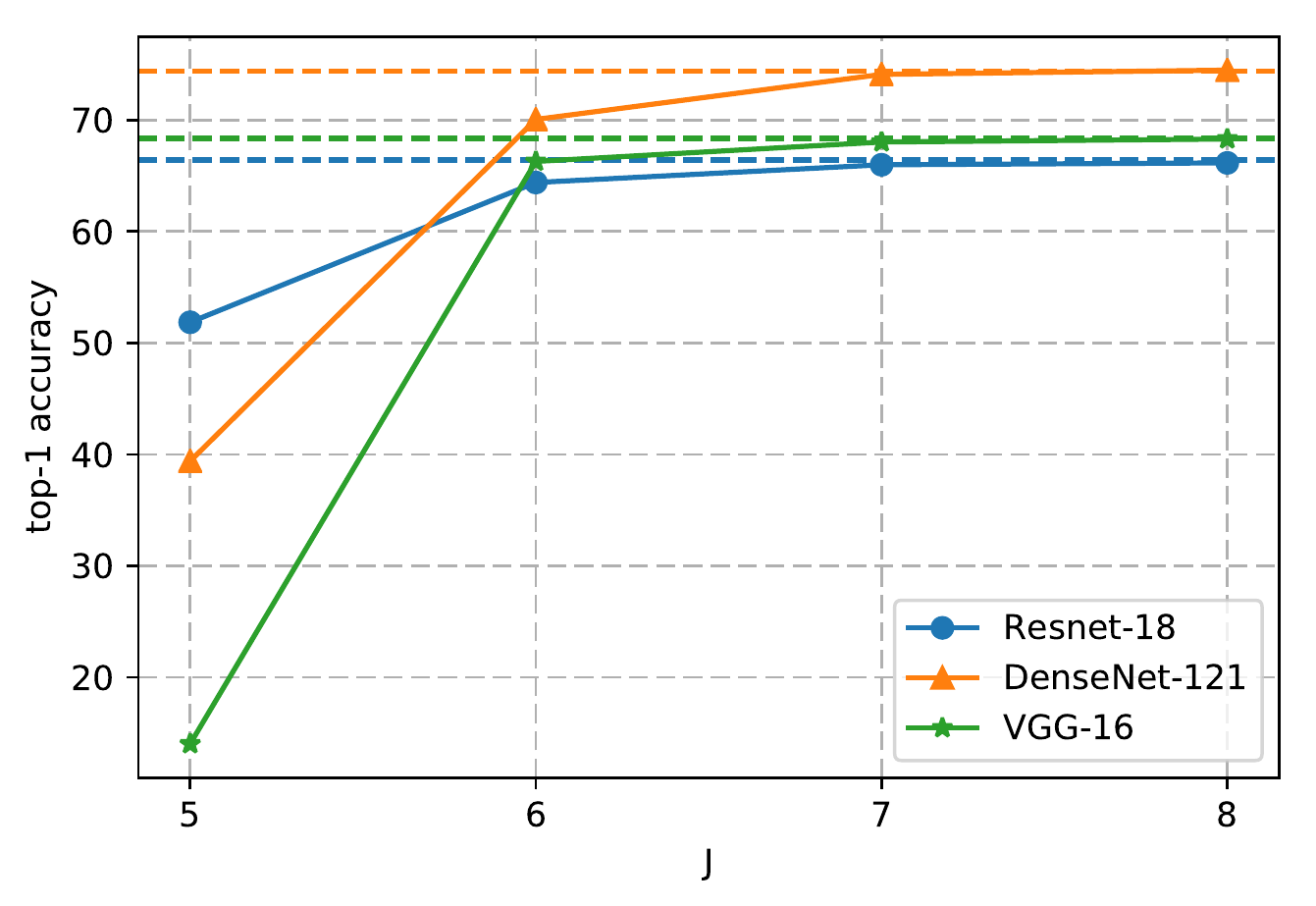}}%
\subfloat[]{
\includegraphics[width=.5\linewidth]{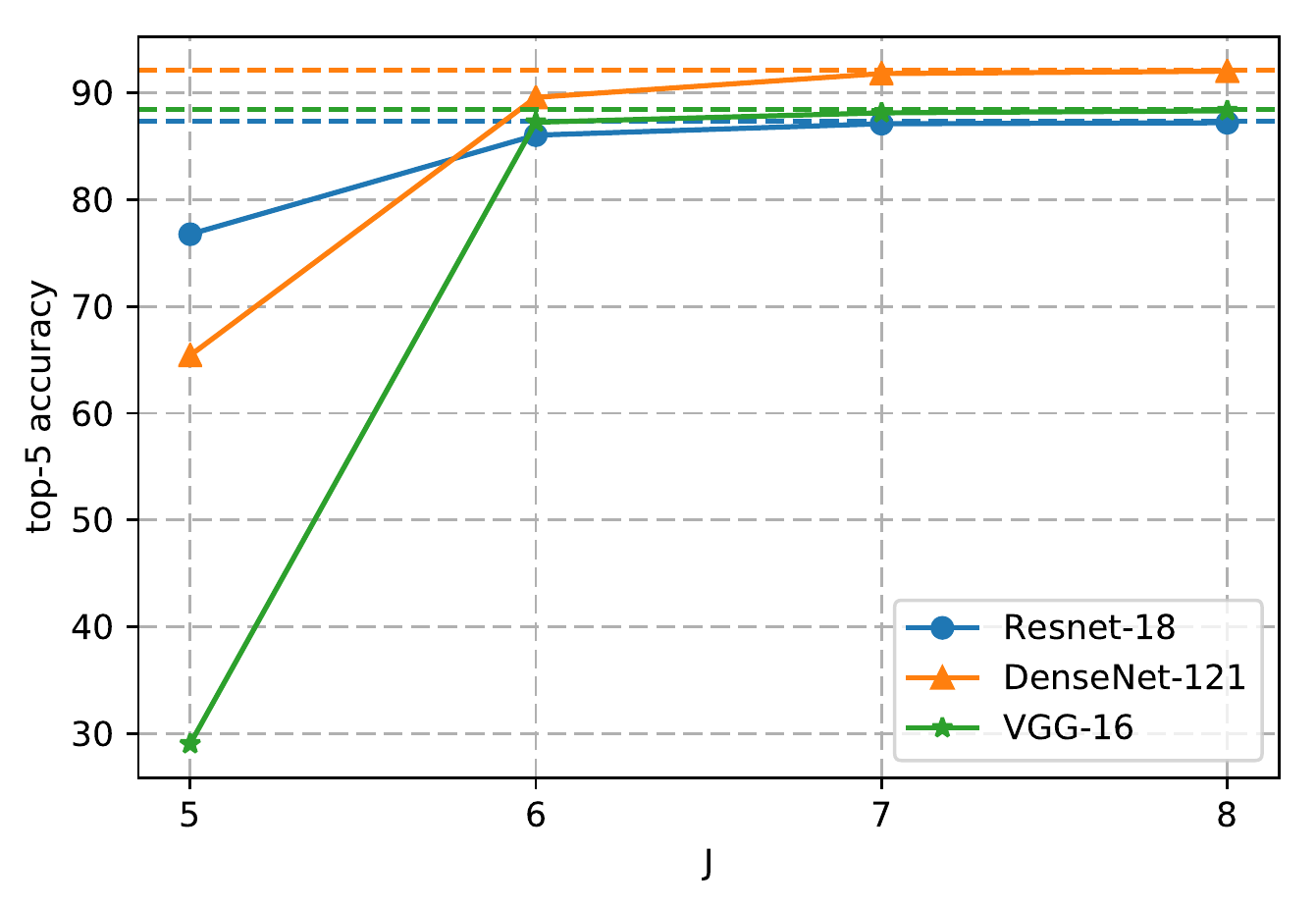}}%
\caption{Performance on ImageNet for different networks with binary tensor expansion using different $J$ bits. Dashed-line indicates FP32 accuracy.
Left is for top-1, right is for top-5. }\label{fig:top1a5}
\end{minipage}
\hspace{1ex}
\begin{minipage}{.495\linewidth}
\centering
\small
\subfloat[]{\label{fig:wd}%
\includegraphics[width=.5\linewidth]{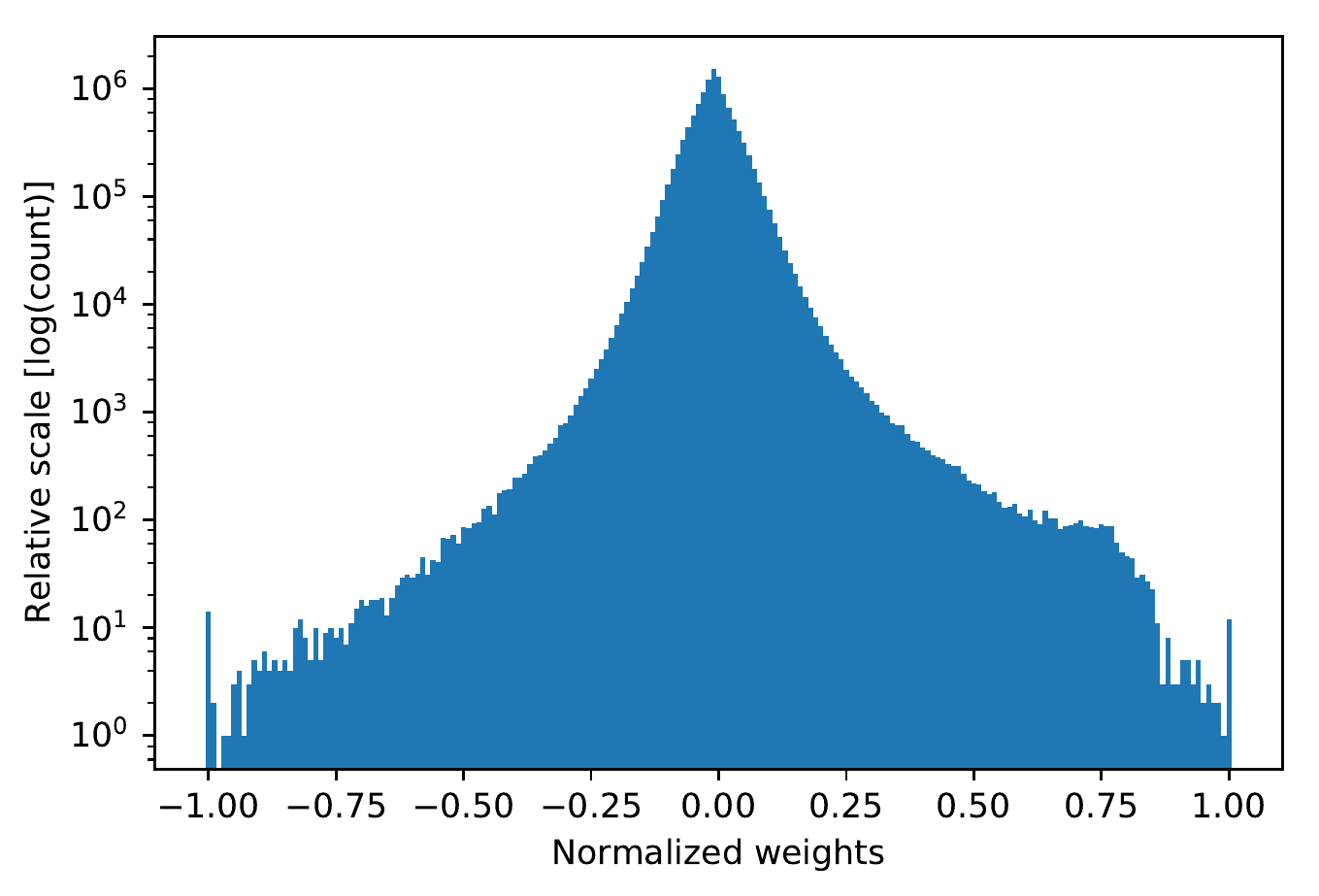}}%
\subfloat[]{\label{fig:sparse}%
\includegraphics[width=.5\linewidth]{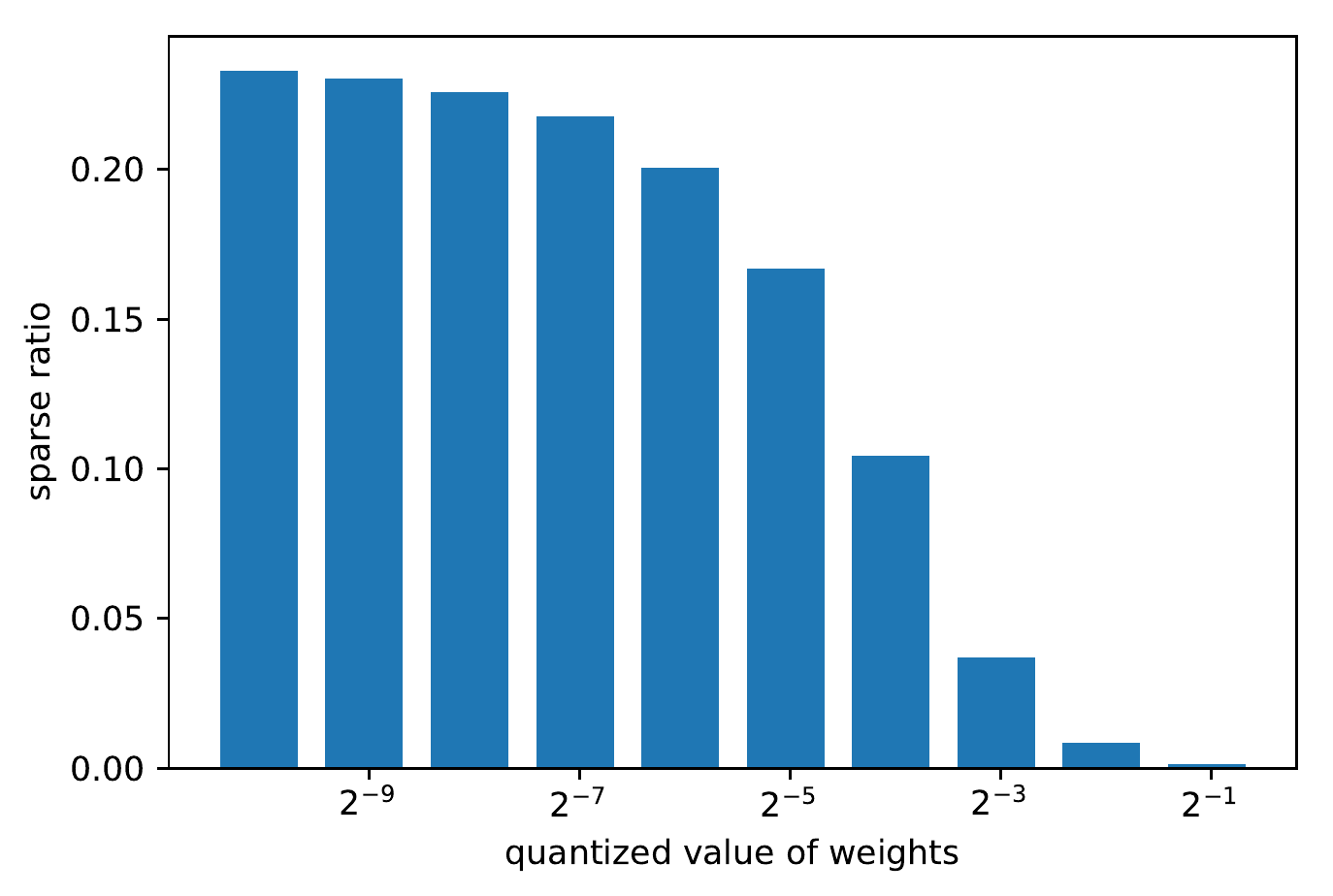}}%
\caption{Weight distribution of all conv-layers for ResNet-18. (a) normalized weights, (b) sparse-ratio of each binary quantized weight.}\label{fig:3}
\end{minipage}
\end{figure*}
\subsection{Binary Network Decomposition}
Few existing works like us consider transferring full-precision networks into multi-bits binary networks in a training-free way.
Binary weighted decomposition (BWD) \cite{kamiya2017binary} takes each filter as a basic unit as BWN, and expands each filter into a linear combination of binary filters and a FP32 scalar.
%
ABC-Net \cite{lin2017abc} approximates full-precision tensor with a linear combination of multiple binary tensors and FP32 scalar weights during training-procedure to obtain multi-bits binary networks. 
Our method is quite different to these two works. We further consider the redundance and sparsity in the expanded binary tensors,
and try to decompose binary tensors. The decomposition is similar to spatial decomposition \cite{Jaderberg2014Speeding} but in the binary space.
Hence, our binary decomposition step can also be viewed as binary spatial decomposition.

\section{Method}
As is known, parameters of each conv-layer in CNNs could be represented as a 4-dimensional (4D) tensor.
We take tensor as our study target. We first present how to expand full-precision tensors into a limited number of binary tensors.
Then we show some binary tensors that fulfill certain conditions can be decomposed into two low-rank binary tensors, and propose an algorithm for that purpose.

\subsection{Tensor Binary Composition}
Suppose the weight parameters of a conv-layer are represented by a 4D tensor $\mathbf{W}_t \in \mathbb{R}^{n\times k\times  k\times  m}$,  where $n$ is the number of input channels, $m$ is the number of output channels, and $k\times k$ is the convolution kernel size.
For each element $w \in {\mathbf{W}}_t$, we first normalize it with
\begin{equation} \label{eq1}
\small
    \tilde {w} = w /w_{max},
\end{equation}
where $w_{max} = \max\nolimits_{w_i\in \mathbf{W}_t }\{\vert w_i \vert\}$. The normalized tensor is denoted as $\tilde {\mathbf{W}}_t$.
The normalization makes every element $\tilde {w}$ within range $[-1, 1]$.
For simplicity, we denote the magnitude of $\tilde{w}$ as $\hat{w}$, i.e., $\tilde{w}$ = $sign(\tilde{w}) \hat{w}$, where $sign(\cdot)$ is a sign function which equals to -1 when $\tilde{w}$ $<$ 0, otherwise 1.
And $\hat{w}$ $\in$ $[0, 1]$ can be expressed by the composite of a series of fixed-point basis as
\begin{equation}
\small
    \hat{w} \approx \sum\nolimits_{i=0}^{J - 2}a_i*2^{-i},
\end{equation}
where $a_i$ $\in$ $\{0, 1\}$ is a binary coefficient indicating whether certain power-of-two term is activated or not, $i \in \{0, \cdots, J-2\}$ means totally $J-1$ bits is needed for the representation. When taking the sign bit into consideration, $\tilde{w}$ requires $J$ bits to represent.

Denote the magnitude of the normalized tensor as $\hat{\mathbf{W}}_t$.
Tensor binary composition is a kind of tensor expansion, when each element of the tensor is binary expanded and expressed by the same bit rate $J$ as
\begin{equation}\label{eq6}
\small
     \hat{\mathbf{W}}_t \approx \sum\nolimits_{i = 0}^{J - 2}A_i*2^{-i},
\end{equation}
where $A_i \in \{0,1\}^{n\times k\times  k\times  m}$ is 4D binary tensor.
$J$ will impact the approximation accuracy, while larger $J$ gives more accurate results.
We empirically study three different ImageNet CNN models. Figure \ref{fig:top1a5} shows that $J = 7$ is already sufficiently good to keep a balance between the accuracy and the efficiency of the expansion.

Different $A_i$ may have different sparsity, which could be further utilized to compress the binary tensor.
Figure \ref{fig:wd} illustrates the distribution of normalized weights $\tilde {w}$ in all the layers of ResNet-18, which looks like a Laplacian distribution, where most weight values concentrate in the range $(-0.5, 0.5)$.
This suggests that 1 is very rare in some binary tensor $A_i$ with smaller $i$, since smaller $i$ corresponds to bigger values in the power-of-two expansion.
Figure \ref{fig:sparse} further shows the average sparsity of each binary tensor $A_i$,
which also verifies that $A_i$ with smaller $i$ is much more sparse. 
Due to the sparsity of $A_i$, we next perform binary tensor decomposition to further reduce the computation complexity, as introduced in the next section.

\subsubsection{Binary expansion with $\alpha$ scaling factor}
The non-saturation direct expansion from FP32 to low-bits will yield non-negligible accuracy loss as shown in \cite{tensorrt,googlewhitepaper}.
A scaling factor is usually introduced and learnt to minimize the loss through an additional calibration procedure \cite{tensorrt,googlewhitepaper}.
Similarly, we impose a scaling factor $\alpha$ to Eq.\eqref{eq1} as
\begin{equation}
\small
    \tilde{w} = \alpha * w /w_{max},
\end{equation}
where $\alpha \geq 1$ is a parameter to control the range of $\tilde {w} \in [-\alpha, \alpha]$.
When the scaling factor $\alpha$ is allowed, the normalized weight $\hat{w} \in [0, \alpha]$ can be expressed with a composite of power-of-two terms as below:
\begin{equation}
\small
    \hat{w} \approx \sum\nolimits_{i = -q}^{J - q - 2}a_i*2^{-i},
\end{equation}
where $q = \lceil \log_2{\alpha} \rceil$ and $J$ also denotes the number bits of the weight, including $J - 1$ bits for magnitude and 1 sign bit.
The corresponding tensor form can be written as
\begin{equation}\label{BDWa}
\small
    \hat{\mathbf{W}}_t^{\alpha} \approx \sum\nolimits_{i = -q}^{J - q - 2}A_i*2^{-i}.
\end{equation}
Note that the scaling factor $\alpha$ will shift the power-of-two bases from $\{2^0, 2^{-1},\cdots, 2^{-J+2}\}$ for Eq.\eqref{eq6} to $\{2^q, \cdots, 2^0, \cdots, 2^{-J+q+2}\}$ for Eq.\eqref{BDWa}.
When $\alpha = 1$, we have $q = \lceil \log_2{\alpha} \rceil = 0$, which makes Eq.\eqref{BDWa} reduce to the case without scaling factor as in Eq.\eqref{eq6}.

\subsection{Binary Tensor Decomposition}
We have shown that some binary tensors $A_i$ are sparse. As sparse operations require specific hardware/software accelerators, it is not preferred in many practical usages.
In deploy-time network optimization, researches show that full-precision tensor could be factorized into two much smaller and more efficient tensors \cite{Jaderberg2014Speeding,Zhang2016Accelerating}.
Here, we attempt to extend the spatial-decomposition \cite{Jaderberg2014Speeding} to our binary case.

For the simplicity of analysis, we flatten the 4D tensor $\hat{\mathbf{W}}_t \in \mathbb{R}^{n\times k\times k\times m}$ into the weight matrix $\mathbf{W} \in \mathbb{R}^{(n\times k)\times(k\times m)}$ ,
so does for each $A_i$. Here the matrix height and width are $n\times k$ and $k\times m$ respectively. We then factorize a sparse matrix $A$ into two smaller matrices as
\begin{equation}\label{eq7}
\small
    A = B*C,
\end{equation}
where matrix $B \in \{0, 1\}^{(n\times k)\times c}$ and matrix $C \in \{0,1\}^{c\times (k\times m)}$. Note that our method is significantly different from the vector decomposition method~\cite{kamiya2017binary}, which keeps $B$ binary, the other full-precision.
On the contrary, we keep both $B$ and $C$ binary.
This decomposition has the special meaning in conv-layers. It decomposes a conv-layer with $k\times k$ spatial filters into two layers---one layer with $k\times 1$ spatial filters and the other with $1 \times k$ spatial filters.
Suppose the feature map size is $h\times w$, then the number of operations is $n\times m\times k^2 \times h\times w$ for matrix $A$,
while the number of operations reduces to $(m + n)\times c \times k \times h\times w$ for $B * C$.
We have the following lemma regarding to the difference before and after binary decomposition.
\begin{lemma}\label{lemma1}
(1) The computing cost ratio for $A$ over $B * C$ is $\nicefrac{n\times m\times k}{c\times (m+n)}$.
(2) The bit-rate compression ratio from $A$ to $B * C$ is also $\nicefrac{n\times m\times k}{c\times (m+n)}$.
(3) $c < \frac{n\times m\times k}{(m+n)}$ can yield real parameter and computing operation reduction.
\end{lemma}

\subsubsection{Binary Matrix Decomposition}
We first review the property of matrix rank:
\begin{equation}
\small
    rank(B*C)  \leq min\{rank(B),rank(C)\}.
\end{equation}
Comparing with binary matrix factorization methods~\cite{zhang2007binary,miettinen2010sparse}, which tend to minimize certain kind of loss like $\vert A - B*C \vert$ and find matrices $B$ and $C$ iteratively,
we attempt to decompose $A$ into matrices $B$ and $C$ without any loss when $c \geq rank(A)$ is satisfied.
\begin{theorem}\label{th1}
If $c \geq rank(A)$, binary matrix $A\in \{0,1\}^{(n\times k)\times(k\times m)}$ can be losslessly factorized into binary matrices
$B \in \{0,1\}^{(n\times k)\times c}$ and $C \in \{0,1\}^{c \times (k \times m)}$.
\end{theorem}
\textbf{Proof}
According to the Gaussian elimination method, matrix $A$ can be converted to an upper triangular matrix $D$. 
Our intuition is to construct matrices $B$ and $C$ through the process of Gaussian elimination. 
Assume $n \leq m$, $P_i$ is the transform matrix representing the $i$-th primary transformation, matrix $D$ can be expressed as:
\begin{equation}
\small
    D = \prod_{i=0}^{k} P_{k-i} \otimes A,
\end{equation}
where $\otimes$ is element-wise binary multiply operator so that
\[
\small
    A\otimes B = (A*B)\mod 2.
\]
For simplicity, we use $*$ instead of $\otimes$ here. As $P_i \in \{0, 1\}^{(n\times k)\times (n\times k)}$ is the permutation transform matrix, the inverse matrix of $P_i$ exists. Therefore, $A$ can be decomposed into the following form:
\begin{equation}
\small
    A = \prod_{i=0}^{k} P_i^{-1} * D
\end{equation}
Since $D$ only contains value $1$ in the first $r$ rows where $r$ = $rank(A)$, $D$ can be decomposed into two matrices:
\begin{equation}
\small
    D = \left[ \begin{array} {c} D_1 \\ 0 \end{array}\right] = \left[ \begin{array}{c} I \\ 0 \end{array} \right] * \left[ \begin{array}{c}
         D_1
    \end{array} \right]
\end{equation}
where $D_1$ $\in$ $\{0, 1\}^{r \times (k\times m)}$ is the first $r$ rows of matrix $D$, $I$ is a $r$$\times$$r$ identity matrix.
Then matrix A can be written as:
\begin{equation}
\small
    A = (\prod_{i=0}^k P_i^{-1} * \left[ \begin{array}{c} I \\ 0 \end{array} \right]) * \left[ \begin{array}{c}
         D_1
    \end{array} \right].
\end{equation}
We then obtain the size $(n\times k)\times r$ matrix $B$ as $B = \prod_{i=0}^k P_i^{-1} * \left[ \begin{array}{c} I \\ 0 \end{array} \right]$, and the size $r \times (k\times m)$ matrix $C$ as $C = D_1$ exactly without any loss.
This procedure also indicates that the minimum bottleneck parameter $c$ is $rank(A)$, i.e., $c \geq rank(A)$.
This ends the proof. $\Box$

Based on this proof, we outline the binary matrix decomposition procedure in Algorithm~\ref{algo-fact}.
From the proof, we should also point out that \textit{the proposed binary decomposition is suitable for both conv-layers and FC-layers}.
Note that for binary permutation matrix $P$, its inverse matrix $P^{-1}$ equals to itself, i.e., $P^{-1} = P$.
Suppose $h_A$ and $w_A$ are height and width of matrix $A$, the computing complexity of $B*P$ is just $O(h_A)$ as $P$ is a permutation matrix.
The computing complexity of Algorithm~\ref{algo-fact} is $O(w_A\times h_A^2)$.
\subsubsection{Losslessly Compressible of $A_i$?}
Theorem~\ref{th1} shows that only when $c \geq rank(A)$, our method could produce lossless binary decomposition, while Lemma~\ref{lemma1} shows that only $c < \frac{n\times m\times k}{(m+n)}$ could yield practical parameter and computing operations reduction. We have the following corollary:
\begin{corollary}\label{cor1}
    Binary matrix $A$ is \textbf{losslessly compressible} based on Theorem~\ref{th1} when $rank(A) \leq c < \frac{n\times m\times k}{(m+n)}$.
\end{corollary}
However, it is unknown which $A_i$ in Eq.\eqref{eq6} was \textit{losslessly compressible} before decomposition.
The brute-force way is trying to decompose each $A_i$ as in Eq.\eqref{eq7}, and then keeping those satisfying Definition~\ref{cor1}.
This is obviously inefficient and impracticable.
\setlength{\textfloatsep}{1ex}
\begin{algorithm}[t]
\begin{scriptsize}
\caption{Binary matrix decomposition}\label{algo-fact}
\hspace*{\algorithmicindent} \textbf{Input:} binary matrix $A$ with size $h_A \times w_A$ \\
\hspace*{\algorithmicindent} \textbf{Output:} matrix rank $r$, matrix $B$, matrix $C$ ($A = B*C$)
\begin{algorithmic}[1]
\Function{BinaryMatDecomposition}{$A$}

\If{$h_A \leq w_A$ }
    \State $A \gets A^T$
    \State $transpose = True$
\EndIf
\State $r \gets 0$;
\State $B \gets$ identity matrix $h_A \times w_A$
\For {$c \gets$ 1 to $w_A$}
    \State $l \gets$ first raw satisfy the constraints: $A[l,c] = 1$ and $l \geq r+1$
    \State Reverse row $l$ \& $r+1$, $P \gets$ corresponding transition matrix
    \State $r \gets r+1$;
    \State $B \gets B*P^{-1}$
    \For {$row \gets r+1$ to $h_A$}
        \If {$A[row, c] > 0$}
            \State $A[row,:] \gets (A[r,:] + A[row,:])$ mod 2
            \State $P \gets$ corresponding transition matrix
            \State $B \gets B*P^{-1}$
        \EndIf
    \EndFor
\EndFor
\State $C \gets$ first $r$ rows of A
\State $P \gets$ first $r$ rows are identity matrix, other $h_A - r$ rows are zeros.
\State $B \gets B*P$
\If{$transpose$}
    \State Return $r$, $C^T$, $B^T$
\Else
    \State Return $r$, $B$, $C$
\EndIf
\EndFunction
\end{algorithmic}
\end{scriptsize}
\end{algorithm}
Alternatively, we may use some heuristic cue to estimate which subset of $\{A_i\}$ could be \textit{losslessly compressible}.
According to Figure \ref{fig:3}, $A_i$ with smaller $i$ is more sparse than that with bigger $i$. Empirically, more sparsity corresponds to smaller $rank(A_i)$.
Based on Theorem~\ref{th1}, this pushes us to seek the watershed value $j$ so that those $A_i$ where $i\leq j$ are \textit{losslessly compressible}, while other $A_i$ ($i>j$) are not.
This requires introducing a variable into the definition of $A_i$, so that we choose the $A_i$ defined by Eq.\eqref{BDWa} rather than Eq.\eqref{eq6}.
As is known, the $A_i$ sequence defined by Eq.\eqref{BDWa} is $\{A_{-q}, \cdots, A_0, \cdots, A_{J-q-2}\}$ where $q = \lceil \log_2{\alpha} \rceil$.
Here, we seek for the optimal $\alpha$, so that $j=0$ is the watershed, i.e., $\{A_{-q}, \cdots, A_0\}$ are \textit{losslessly compressible}.

\if 0
\begin{figure*}[t]
\begin{minipage}{.66\linewidth}
\centering
\small
\subfloat[]{\label{fig:2a}%
\includegraphics[width=.5\linewidth]{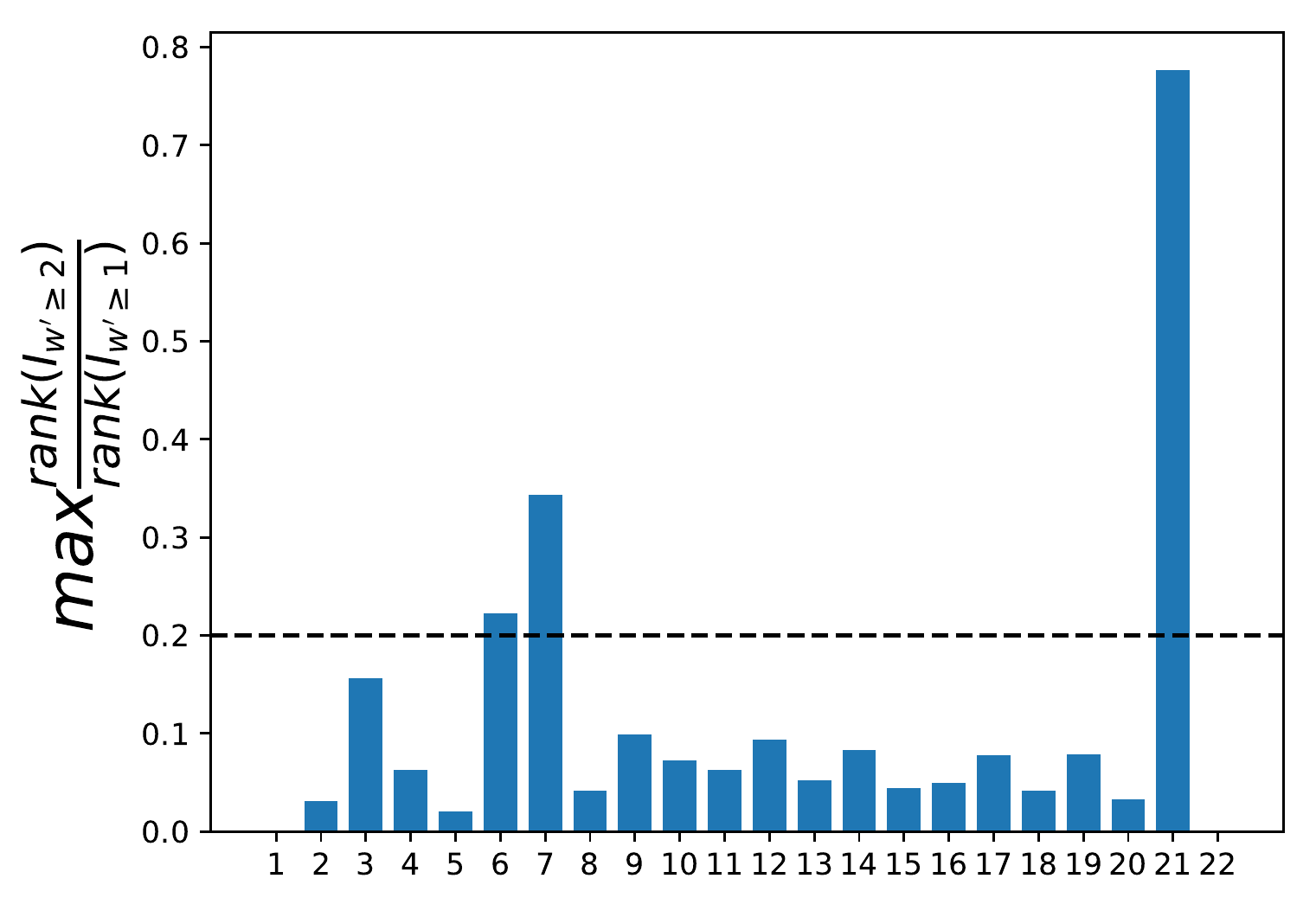}}%
\subfloat[]{\label{fig:3a}%
\includegraphics[width=.5\linewidth]{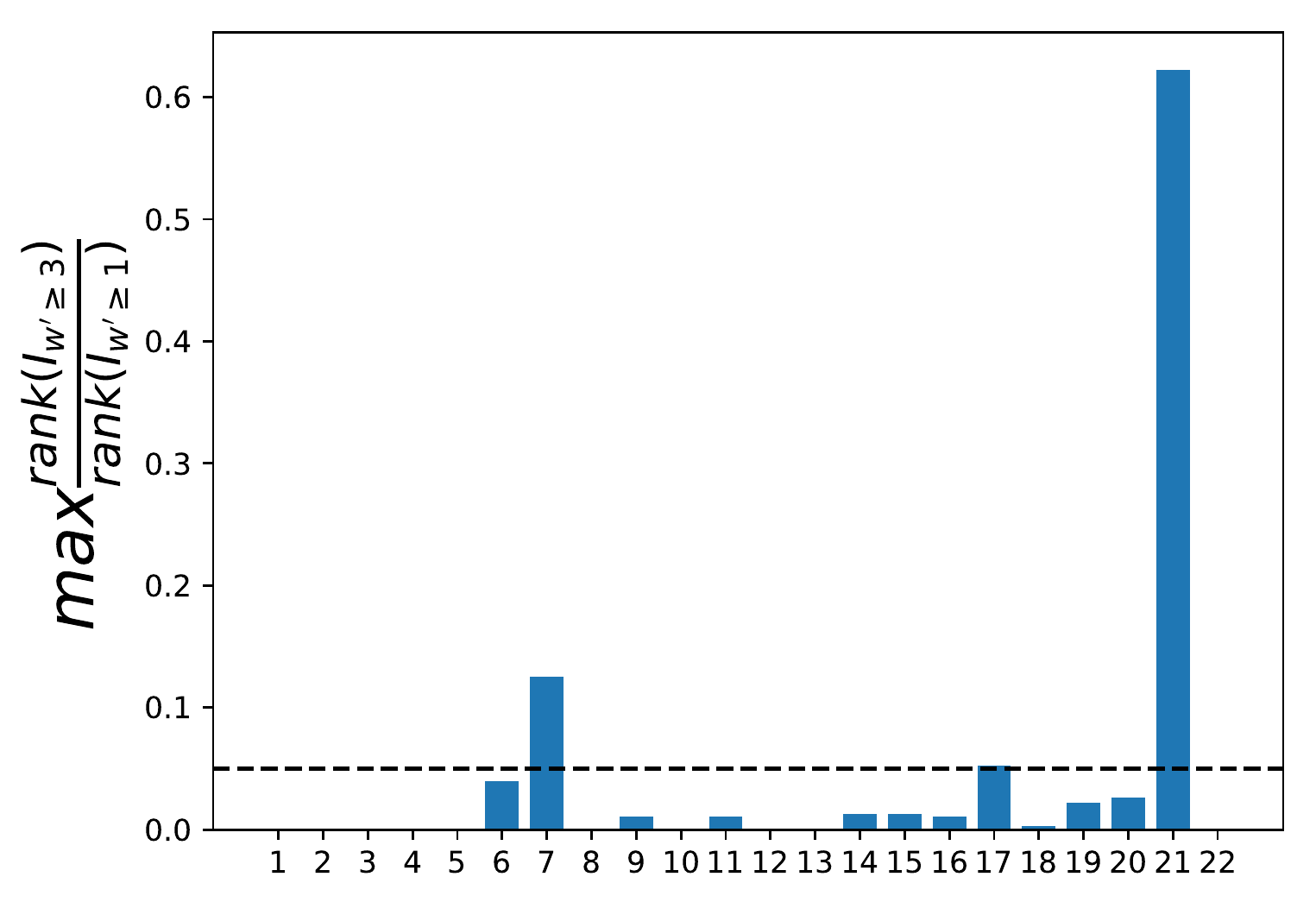}}%
\caption{Maximum rank ratio per-layer of ResNet-18.}
\end{minipage}
\hspace{1ex}
\begin{minipage}{.33\linewidth}
   \centering
   \small
   \includegraphics[width=0.98\linewidth]{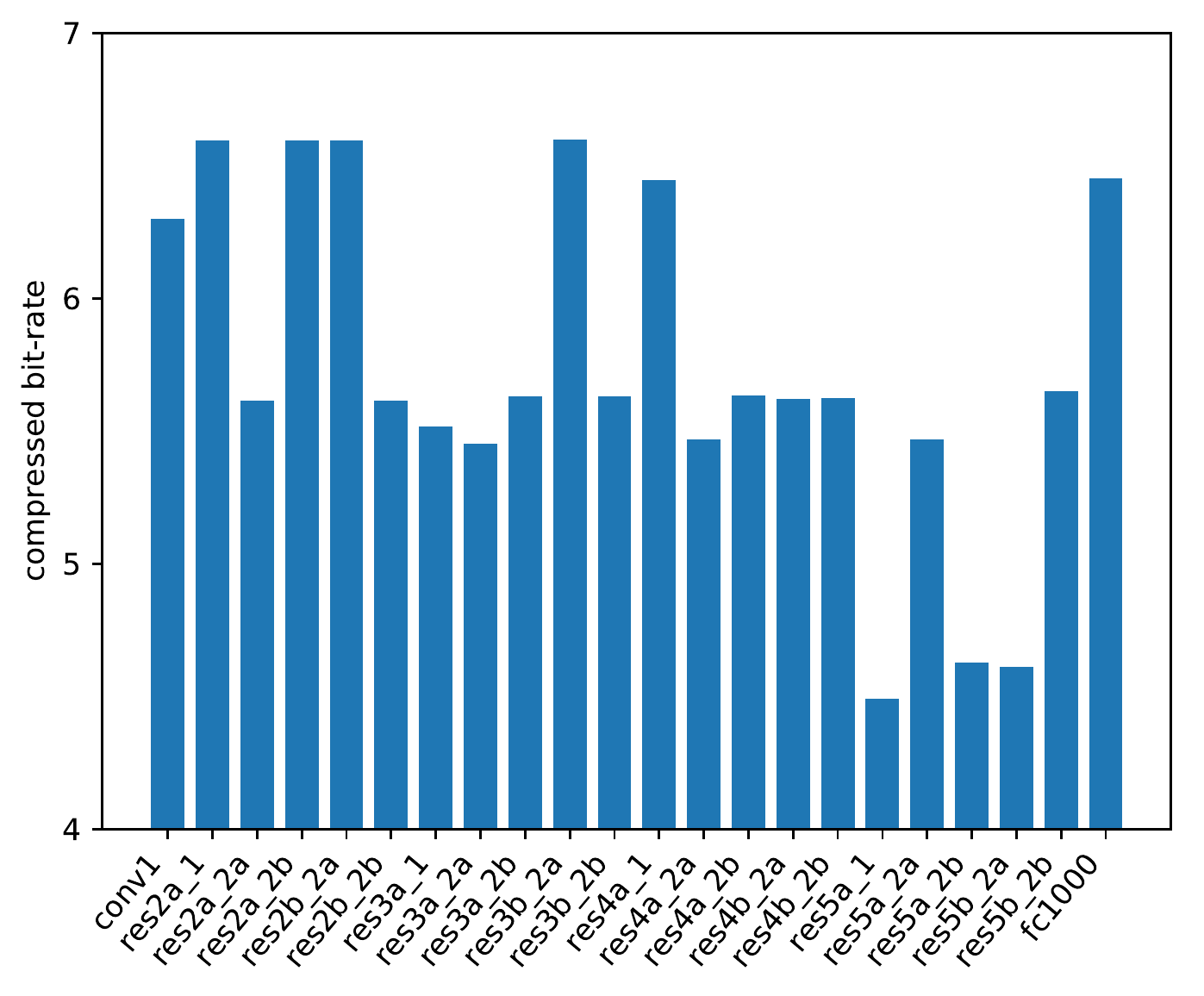}
   \caption{Bit-rate per-layer of ResNet-18.}\label{fig:per}
\end{minipage}
\end{figure*}

\else
\begin{figure}[t]
\centering
\small
\subfloat[]{\label{fig:2a}%
\includegraphics[width=.495\linewidth]{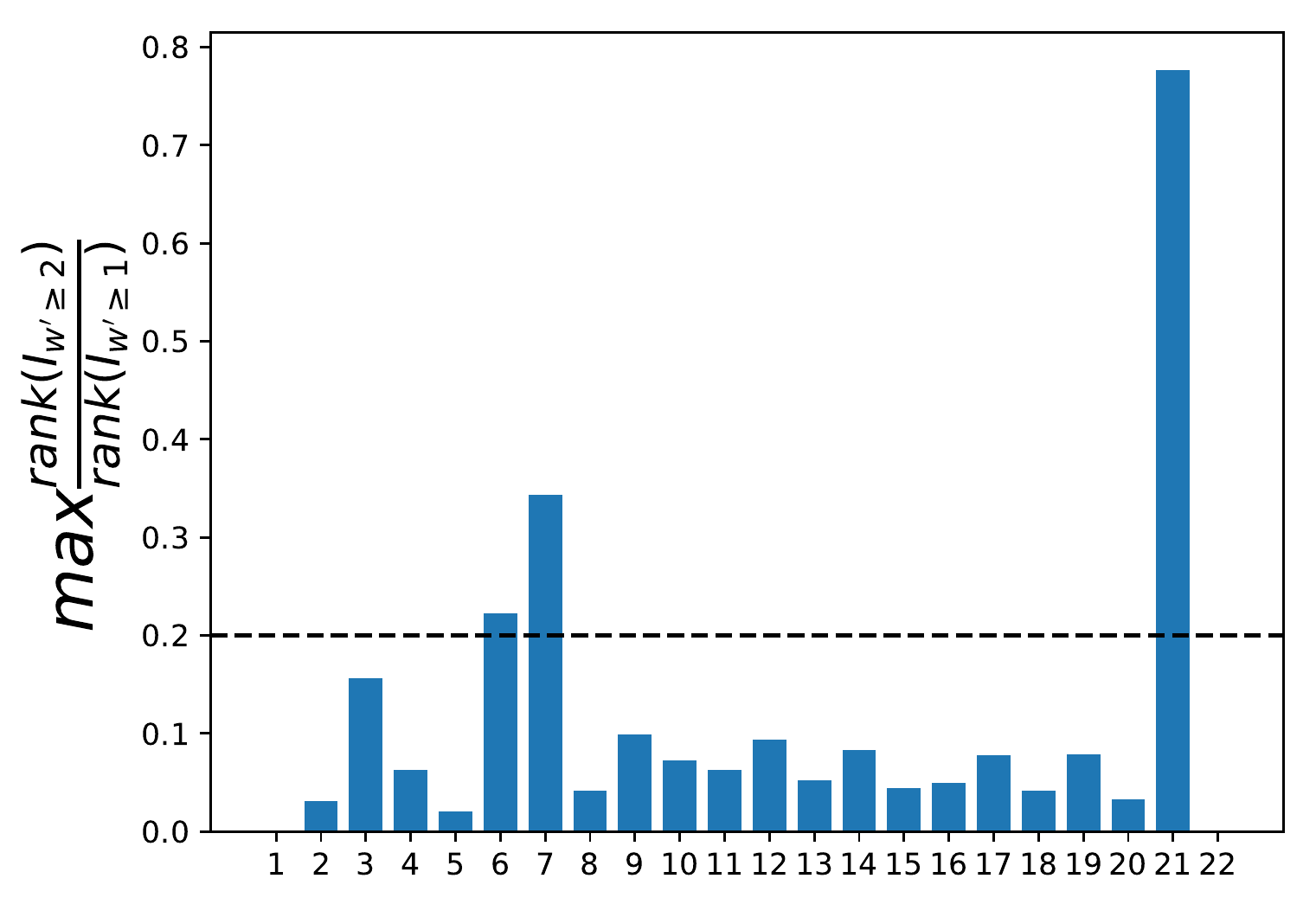}}%
\subfloat[]{\label{fig:3a}%
\includegraphics[width=.495\linewidth]{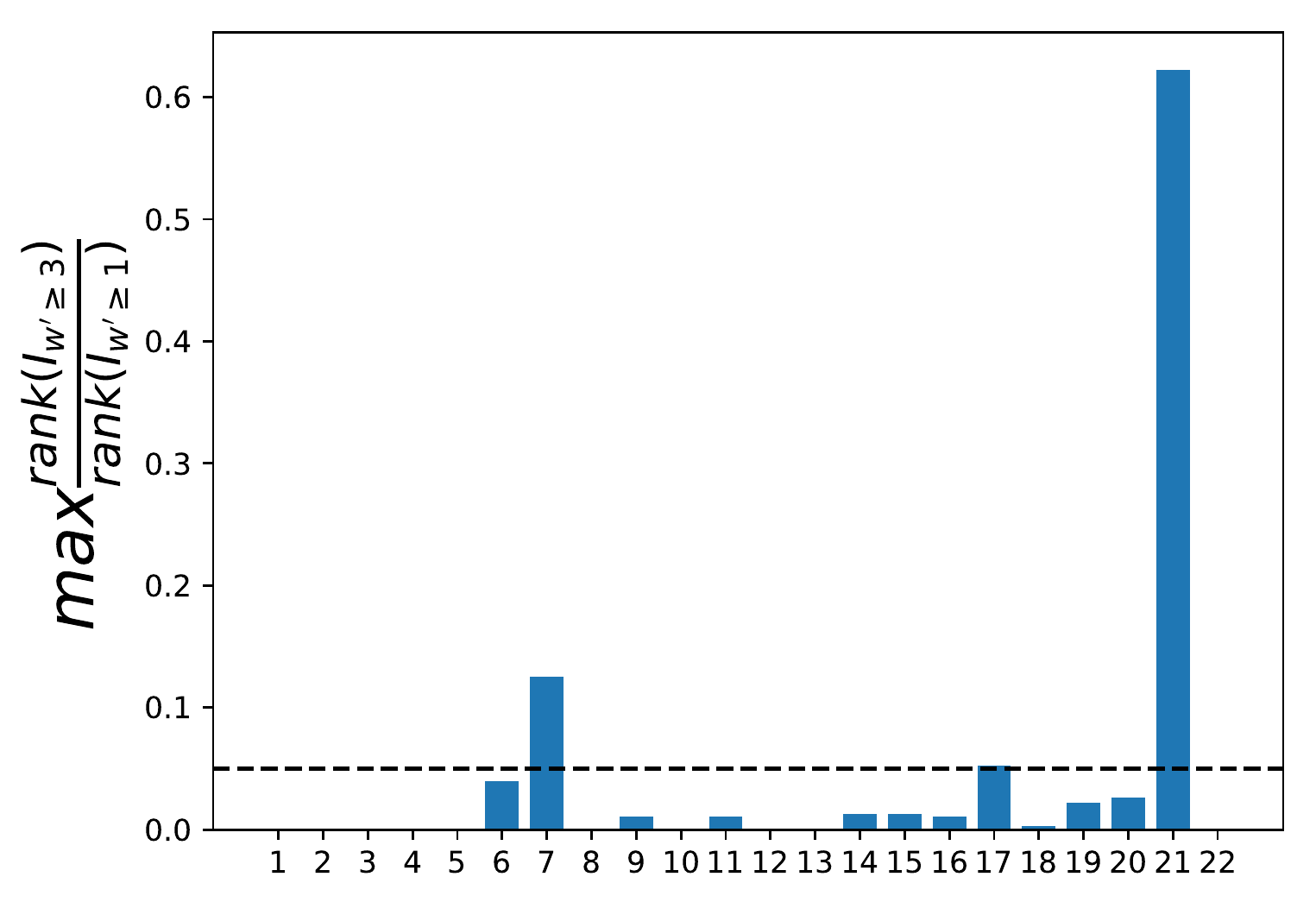}}%
\caption{Maximum rank ratio per-layer of ResNet-18.}
\end{figure}
\fi

For simplicity, we still denote the flatten matrix of the tensor $\hat{\mathbf{W}}_t^{\alpha}$ in Eq.\eqref{BDWa} as $\mathbf{W}$ $\in$ $\mathbb{R} ^{(n\times k)\times (k\times m)}$. We propose to use the indicator matrix described below for easy analysis.
\begin{definition}\label{def1}
The \textbf{indicator matrix} of $\mathbf{W}$ is defined as $I_{w>\beta} \in \{ 0,1 \}^{(n\times k)\times (k\times m)}$, in which the value at position $(x, y)$ is
    $I_{w > \beta}[x, y] = I_f(\mathbf{W}[x,y] > \beta)$,
where $\beta \in [0, \alpha]$ is a parameter, $I_f(\cdot)$ is an element-wise indication function, which equals to 1 when the prediction is true, otherwise 0.
\end{definition}
Based on this definition, $A_i$ in Eq.\eqref{BDWa} can be written as
\begin{equation}\label{Ai}
\small
    A_i[x,y] = I_f(\lfloor \frac{\mathbf{W}[x,y] + \Delta w}{2^{-i}} \rfloor \mod 2 = 1),
\end{equation}
where $\Delta w = 2^{-J + q + 1}$ is the largest throwing-away power-of-two terms in Eq.\eqref{BDWa}.


Define $w' = w + \Delta w$, according to Eq.\eqref{BDWa}, the rank of matrix $A_0$ can be expressed as:
\begin{equation}
\small
\begin{split}
    rank(A_0) &= rank(\sum\nolimits_{i = 0}^{\lceil n_I/2 \rceil -1} I_{(2i+1) \leq w' < (2i+2)}) \nonumber \\
    &= rank(\sum\nolimits_{i=1}^{n_I} I_{w' \geq i}),
\end{split}
\end{equation}
where $n_I$ = $\lceil \alpha \rceil$. Based on the matrix rank property
\begin{equation}\label{A+B}
\small
    rank(A + B) \leq rank(A) + rank(B),
\end{equation}
we derive the upper bound of the $rank(A_i)$ as:
\begin{equation}\label{AiB}
\small
    rank(A_0) \leq \sum\nolimits_{i=1}^{n_I} {rank(I_{w' \geq i})}.
\end{equation}

The empirical results show that when $rank(I_{w' \geq 1}) \leq 0.5 * min\{n\times k, m\times k\}$, and the rank of the indicator matrix satisfies the following constraints:
\begin{equation}
\small
\begin{split}
    &(1)~ \frac{rank(I_{w' \geq 2})}{rank(I_{w' \geq 1})} \leq C_0,  \\
    &(2)~ max\{\frac{rank(I_{w' \geq i})}{rank(I_{w' \geq 1})}\} \leq C_1, 3\leq i \leq n_I.
\end{split}
\end{equation}
With the bound \eqref{AiB}, we get the bound of the rank by the indicator matrix $I_{w'\geq1}$:
\begin{equation}
\small
    rank(A_0) \leq ((\alpha - 2)_+ * C_1 + C_0 + 1)*rank(I_{w' \geq 1}),
\end{equation}
where $(x)_+$ equals to $x$ if $x > 0$, otherwise 0.
We make some empirical study on ResNet-18, and find that $C_0 \leq 0.2$ in most of the layers as in Figure \ref{fig:2a}, $C_1 \leq 0.05$ in most of the layers as in Figure \ref{fig:3a}, and $\alpha \leq 8$ in all the layers.
Hence, we have a simpler method to estimate $rank(A_0)$ by:
\begin{equation}
\small
\begin{split}
 rank(A_0) &\leq (4 * 0.05 + 0.2 + 1) * rank(I_{w' \geq 1}) \nonumber \\
            &=1.4 * rank(I_{w' \geq 1}). \nonumber
\end{split}
\end{equation}
Therefore, we transfer the optimization of $\alpha$ to the optimization of $rank(I_{w' \geq 1})$.

\setlength{\textfloatsep}{1ex}
\begin{algorithm}[t]
\begin{scriptsize}
\caption{Binary search $\alpha$ to satisfy rank condition.}\label{algo-rank}
\hspace*{\algorithmicindent} \textbf{Input:} weight matrix $\mathbf{W}$, expected rank $c$\\
\hspace*{\algorithmicindent} \textbf{Output:} scalar value $\alpha$
\begin{algorithmic}[1]
\Function{ScalarValueSearch}{$W$, $c$}

\State $min \gets 0$, $max \gets$ max number of 1 value in a full rank matrix
\State sort $\mathbf{W}$ in a descending order to a vector $\mathbf{v}$
\While {$min\leq max$}
    \State $center \gets (min+max)/2$
    \State $\alpha \gets 1/\mathbf{v}[center]$
    \State Compute indicator matrix $I_{w\geq 1}$
    \State Compute rank $r$ of $I_{w\geq 1}$
    \If {$r > c$}
        \State $max \gets center-1$
    \ElsIf {$r < c$}
        \State $min \gets center+1$
    \Else
        \State Return $\alpha$
    \EndIf
\EndWhile
\State Return $\alpha = 1/\mathbf{v}[max]$
\EndFunction
\end{algorithmic}
\end{scriptsize}
\end{algorithm}

\subsection{A binary search algorithm for scaling factor $a$}
In practice, we may give an expected upper bound $c$ for $rank(I_{w \geq 1})$, and search the optimal $\alpha$ to satisfy
\begin{equation}
\small
    rank(I_{w \geq 1}) \leq c.
\end{equation}
As $w \in [0, \alpha]$ and $\alpha > 1$, $w \in [0, 1]$ only corresponds to $1/\alpha$ portion of the whole range of $w$.

Generally, weight matrix $\mathbf{W}$ should be sorted and traversed to compute the $rank(I_{w \geq 1})$.
Instead of using this time-consuming method, we propose an efficient solution based on the binary search algorithm.

Suppose that every element in weight matrix $\mathbf{W}$ has a unique value, we sort $\mathbf{W}$ to be a vector $\mathbf{v}$ with length $N = n\times m\times k^2$ in descending order, so that $\mathbf{v}[1]$ is the largest element. Assume index $i$ satisfy the constraint: $\mathbf{v}[i] \geq 1 > \mathbf{v}[i+1]$, then the indicator matrix can be expressed as:
\begin{equation}
\small
     I_{w \geq 1} = I_{p(w) < i+1}
\end{equation}
where $p(w)$ is the index position of the element $w$ in array $\mathbf{v}$, i.e., $\mathbf{v}[p(w)] = w$.
Hence, there are $i$ ones in the indicator matrix while others are zeros.
Comparing matrix $I_{p(w) < i+1}$ with matrix $I_{p(w) < i}$, we have the property:
\begin{equation}
\small
    I_{p(w) < i+1}[x, y] =
    \begin{cases}
            I_{p(w) < i}[x,y]  & p(w[x,y])\neq i \\
            I_{p(w) < i}[x,y] + 1 & p(w[x,y])=i
    \end{cases}
\end{equation}
With the property of Eq.\eqref{A+B}, we get the relation between the indicator matrices with adjacent index:
\begin{equation}
\small
    \vert rank(I_{p(w) < i+1}) - rank(I_{p(w) < i}) \vert \leq 1.
\end{equation}
This indicates that $rank(I_{p(w) < i})$ are continuous integers. Hence, $i \in [i_1, i_2]$ always exists to satisfy below constraints:
\begin{equation}\label{PforI}
\small
\begin{split}
    & rank(I_{p(w) < i})= c , \\
    & (rank(I_{p(w) < i_1}) - c)*(rank(I_{p(w) < i_2}) - c) < 0 .
\end{split}
\end{equation}
Eq.\eqref{PforI} provides the property to design a binary search algorithm as outlined in Algorithm \ref{algo-rank},
which could find the local optimal of $\alpha$ when matrix $\mathbf{W}$ has some identical elements.
The computing complexity of this algorithm is $O(N \log(N))$, where $N = n\times k^2\times m$.

Here, $c$ is a tunable hyper-parameter. In practice, we did not directly give $c$ since different layers may require different $c$.
Instead, we assume $m \geq n$ and define $b = \nicefrac{c}{n\times k}$ as bottleneck ratio, since we reduce the neuron number from $n\times k$ to $c$ by the binary matrix $B$. Then we tune $b$ and make it constant over all the layers, while $b < 0.5$ could provide compression effect.
After getting $\alpha$, we obtain $q = \lceil \log_2{\alpha} \rceil$, and only do binary decomposition for the subset $\{A_{-q}, \cdots, A_0\}$ as they are \textit{losslessly compressible}.

\begin{table}[tp]
\centering
\small
\resizebox{1.0\linewidth}{!}{%
\begin{tabular}{c|c|c|c|c|c|c|c}
\hline
\multirow{2}{*}{Model} & \multicolumn{3}{c|}{FP32} & \multicolumn{3}{c|}{CBDNet} & \multirow{2}{*}{Bitrate} \\ \cline{2-7}
                       & top-1(\%)       & top-5(\%) & size-MB      & top-1(\%)        & top-5(\%)        & size-MB                          \\ \hhline{=|=|=|=|=|=|=|=}
ResNet-18              & 66.41       & 87.37 & 44.57       & 65.27(-1.14) & 86.62(-0.75) & 7.31 & 5.25                      \\ \hline
VGG-16                 & 68.36       & 88.44 &527.7       & 67.36(-1.00) & 87.81(-0.63) & 90.3 & 5.47                      \\ \hline
DenseNet-121           & 74.41       & 92.14 & 30.11       & 73.13(-1.28)  & 91.29(-0.85) & 5.38 & 5.72                      \\ \hline
\end{tabular}
}
\caption{Performance of different CNNs on ImageNet and CBDNet. Numbers in bracket indicates accuracy drop relative to FP32 models.
Last-column lists bitrate of CBDNet.}
\label{tb:performance}
\end{table}

\begin{figure*}[t!]
\begin{minipage}{.5\linewidth}
\centering
\small
\subfloat[]{\label{fig:res_1}%
\includegraphics[width=.5\linewidth]{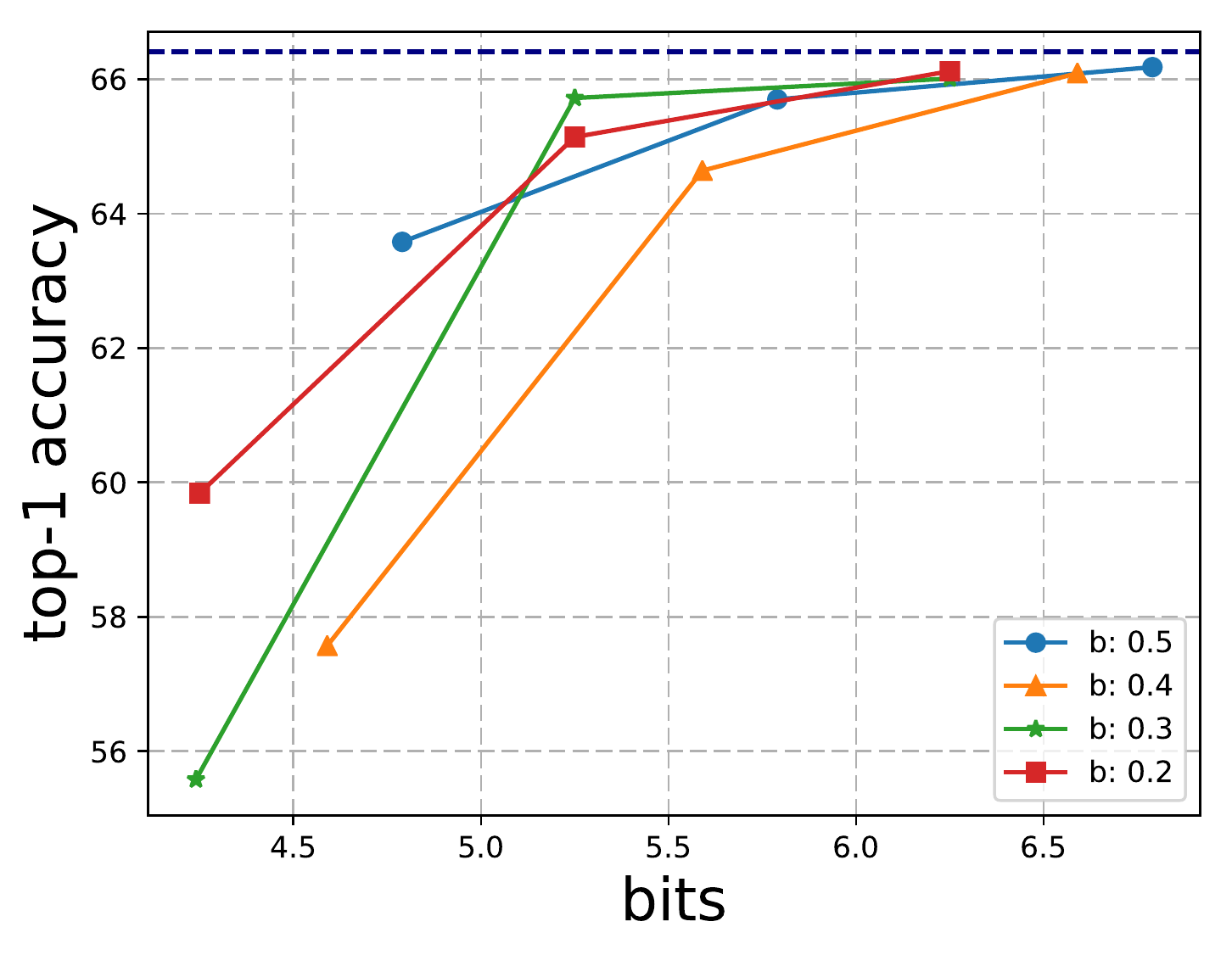}}%
\subfloat[]{\label{fig:res_5}%
\includegraphics[width=.5\linewidth]{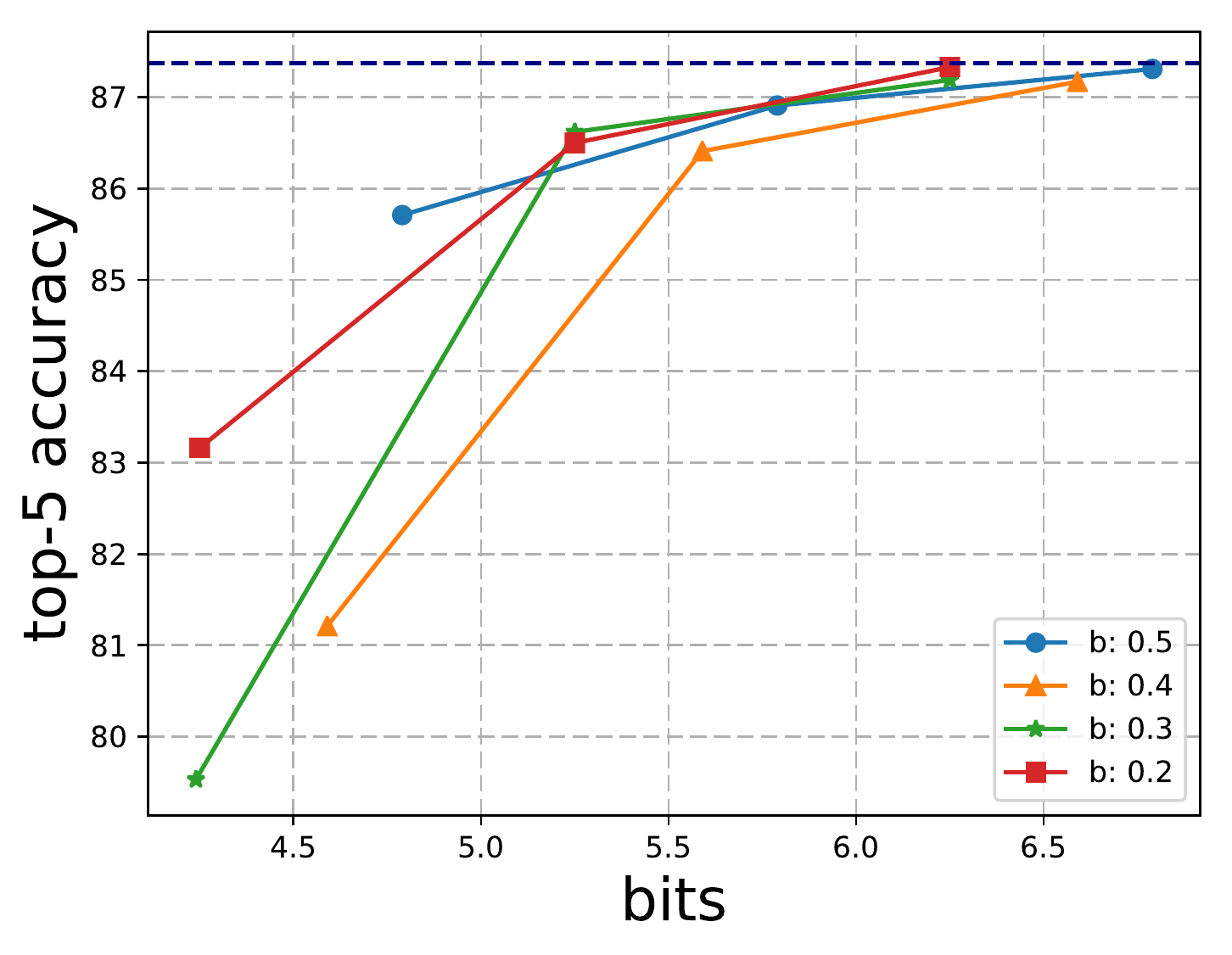}}%
\caption{CBDNet accuracy on ImageNet for ResNet-18. Dashed-line indicates FP32 accuracy. `$b$' for bottleneck ratio.}\label{fig:res}
\end{minipage}
\quad
\begin{minipage}{.5\linewidth}
\centering
\small
\subfloat[]{\label{fig:vgg_1}%
\includegraphics[width=.5\linewidth]{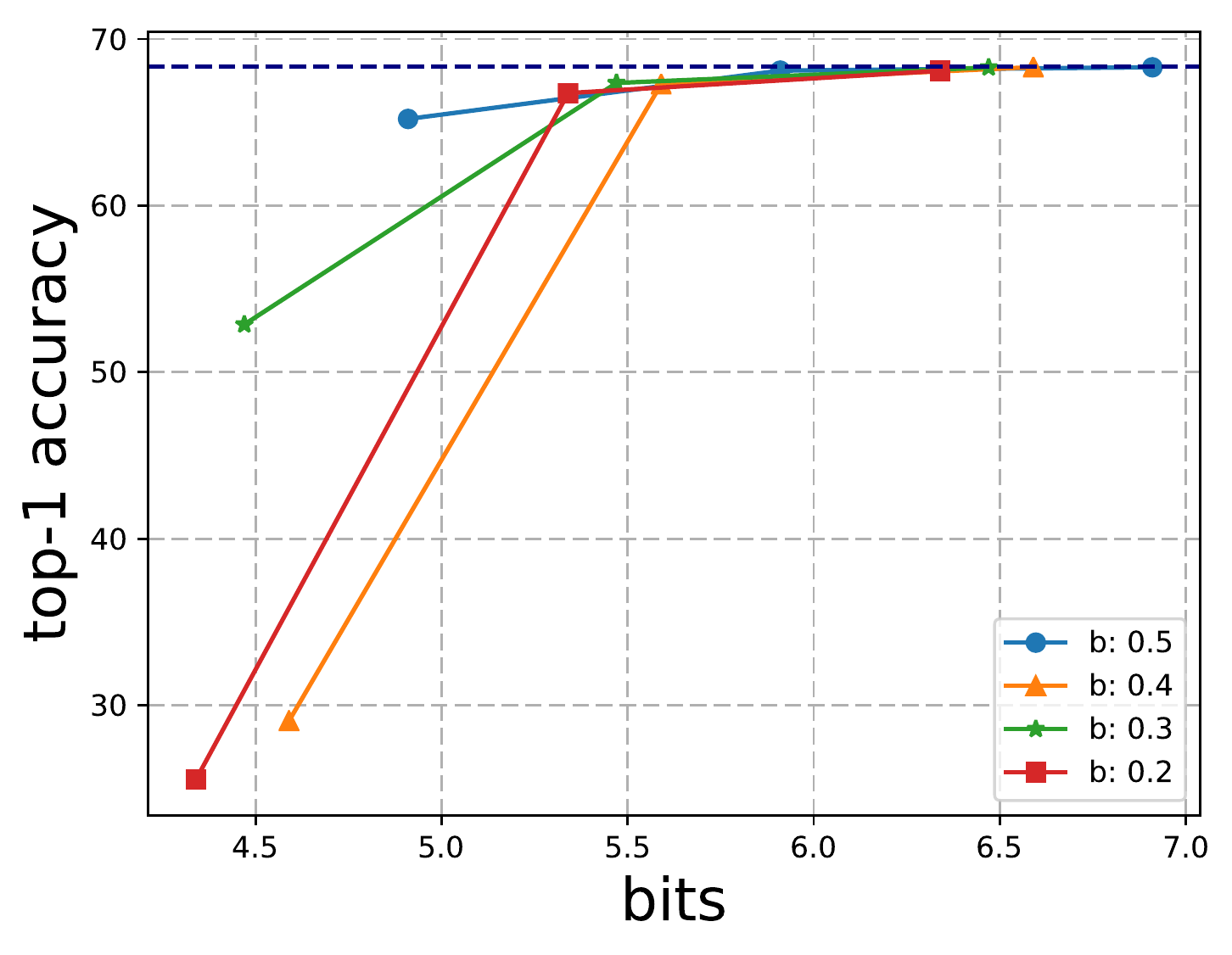}}%
\subfloat[]{\label{fig:vgg_5}%
\includegraphics[width=.5\linewidth]{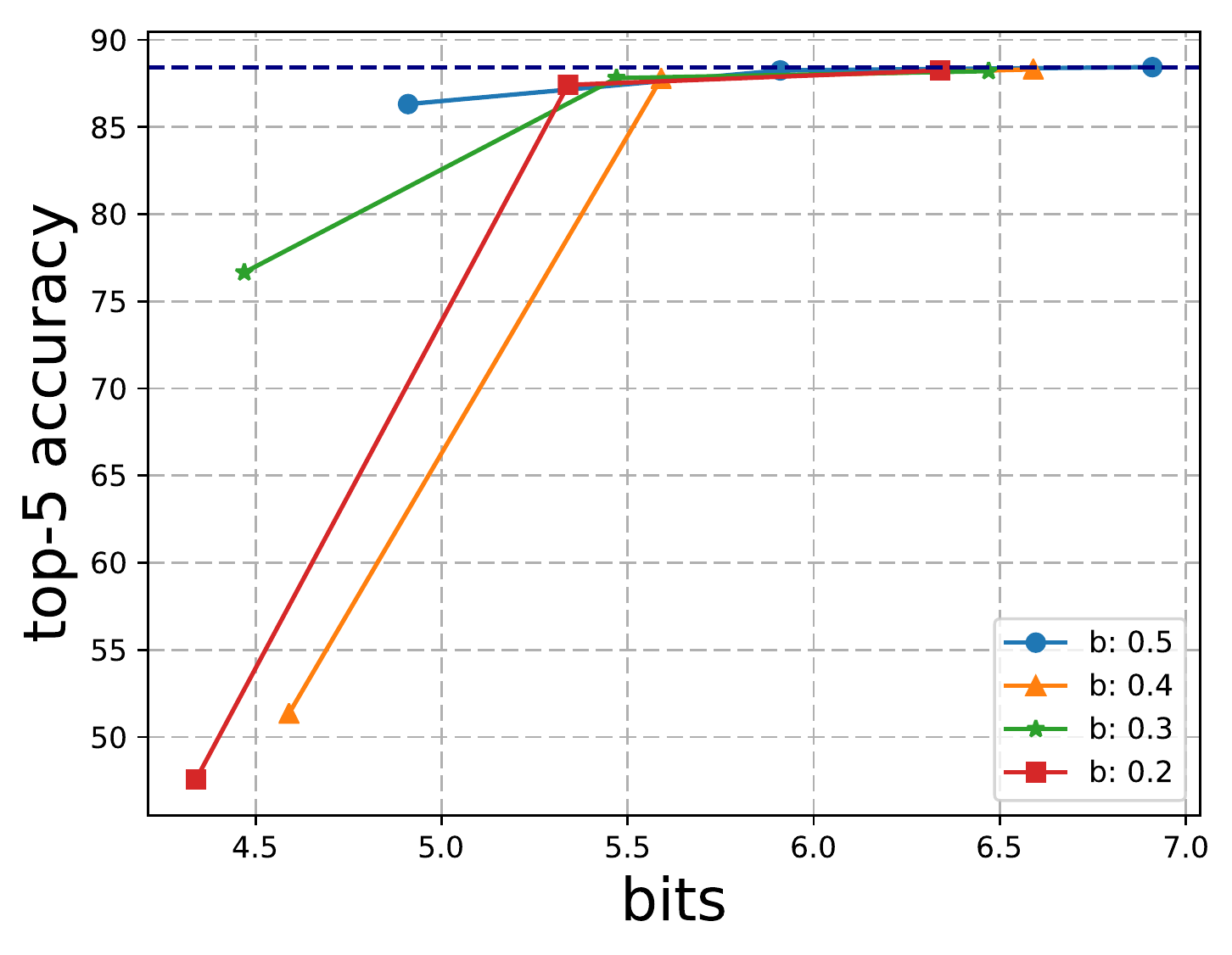}}%
\caption{CBDNet accuracy on ImageNet for VGG-16. Dashed-line indicates FP32 accuracy. `$b$' for bottleneck ratio.}\label{fig:vgg}
\end{minipage}
\end{figure*}

\section{Experiments}
This section conducts experiments to verify the effectiveness of CBDNet on various ImageNet classification networks \cite{ILSVRC15}, as well as
object detection network SSD300 \cite{Liu2016SSD} and semantic segmentation network SegNet \cite{badrinarayanan2017segnet}.

\subsection{Classification Networks on ImageNet}
The ImageNet dataset contains 1.2 million training images, 100k test images, and 50k validation images. Each image is classified into one of 1000 object categories. The images are cropped to $224\times 224$ before fed into the networks. We use the validation set for evaluation and report the classification performance via top-1 and top-5 accuracies. Table \ref{tb:performance} gives an overall performance of different evaluated networks.
Note we decomposed all the conv-layers and FC-layers with the proposed method in this study.
The resulted bit-rate is defined by $32\times size(\mathrm{CBDNet})/size(\mathrm{FP32})$, where $size(x)$ gives the model size of model $x$.

\subsubsection{Model 1:}
\textbf{ResNet-18} \cite{he2015deep} consists of 21 conv-layers and 1 FC-layer. The top-1/top-5 accuracy of the FP32 model is $66.41/87.37$ in our single center crop evaluation.
Figure \ref{fig:res} shows the accuracy and the bit-rate of CBDNet at different bottleneck ratios.
The effective bit-rate of CBDNet is 5.25 with $b$=$0.3$ and $J$=$7$, while the top-1/top-5 accuracy drops by 1.14\% and 0.75\% respectively.

\subsubsection{Model 2:}
\textbf{VGG-16} \cite{simonyan2014very} consists of 13 conv-layers and 3 FC-layers. The top-1/top-5 accuracy of the FP32 model is $68.36/88.44$  in our single center crop evaluation.
Figure \ref{fig:vgg} shows the accuracy and the bit-rate of CBDNet at different bottleneck ratios.
The effective bit-rate of CBDNet is 5.47 with $b$=$0.3$ and $J$=$7$, while the top-1/top-5 accuracy drops by 1\% and 0.63\% respectively.

\begin{table}[t]
\centering
\small
\resizebox{.98\linewidth}{!}{%
\begin{tabular}{c|c|c|c|c|c}
\hline
\multirow{2}{*}{Model} & FP32        & \multicolumn{2}{c|}{BWD}         & \multicolumn{2}{c}{CBDNet}    \\ \cline{2-6}
                       & \begin{tabular}[c]{@{}c@{}}top-5(\%) \end{tabular} & \begin{tabular}[c]{@{}c@{}}top-5(\%) \end{tabular} & Bitrate & \begin{tabular}[c]{@{}c@{}}top-5(\%) \end{tabular} & Bitrate \\ \hline \hline
ResNet-152             & 92.11      & 90.25(-1.86)         & 6        & 91.61(-0.5)         & 5.37     \\ \hline
VGG-16                 & 88.44       & 86.28(-2.16)        & 6        & 87.81(-0.63) & 5.47    \\ \hline
\end{tabular}
}
\caption{CBDNet vs BWD on ResNet and VGGNet.}
\label{tab:comp}
\end{table}

\begin{figure*}[t!]
\begin{minipage}{.5\linewidth}
\centering
\small
\subfloat[]{\label{fig:des_1}%
\includegraphics[width=.5\linewidth]{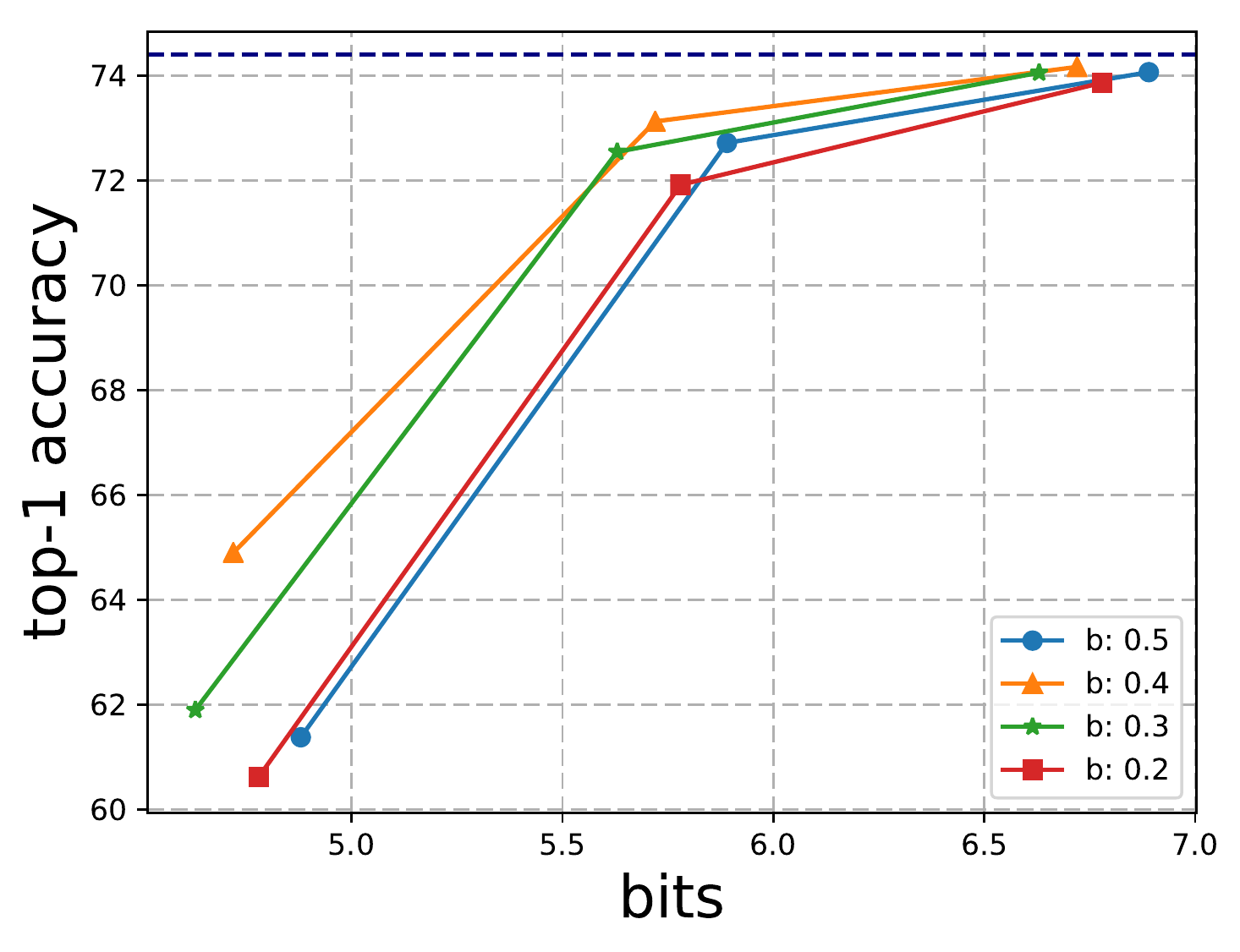}}%
\subfloat[]{\label{fig:des_5}%
\includegraphics[width=.5\linewidth]{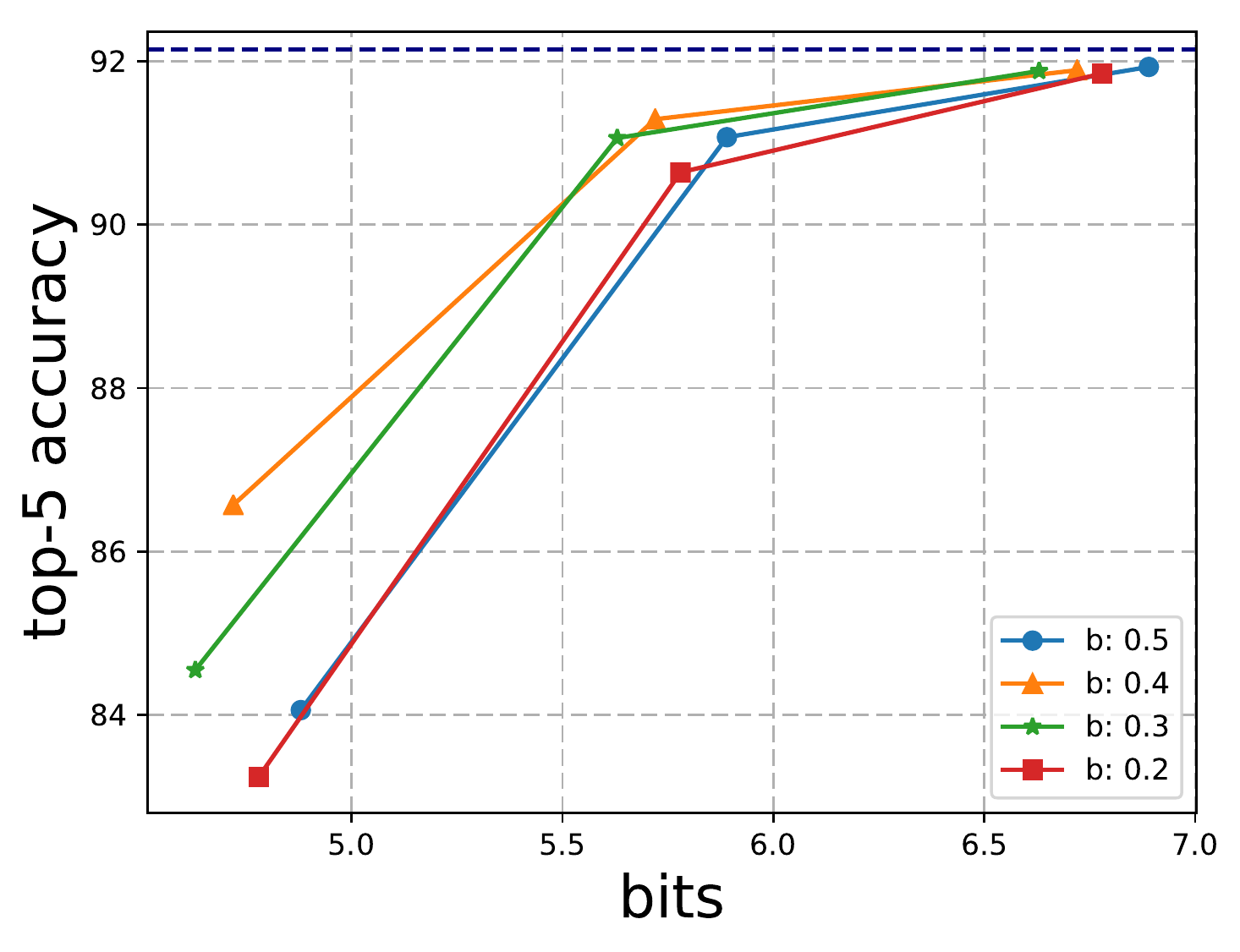}}%
\caption{CBDNet accuracy on ImageNet for DenseNet-121. Dashed-line indicates FP32 accuracy. `$b$' for bottleneck ratio.}\label{fig:des}
\end{minipage}
\quad
\begin{minipage}{.5\linewidth}
\centering
\small
\subfloat[]{\label{fig:ssd}%
\includegraphics[width=.5\linewidth]{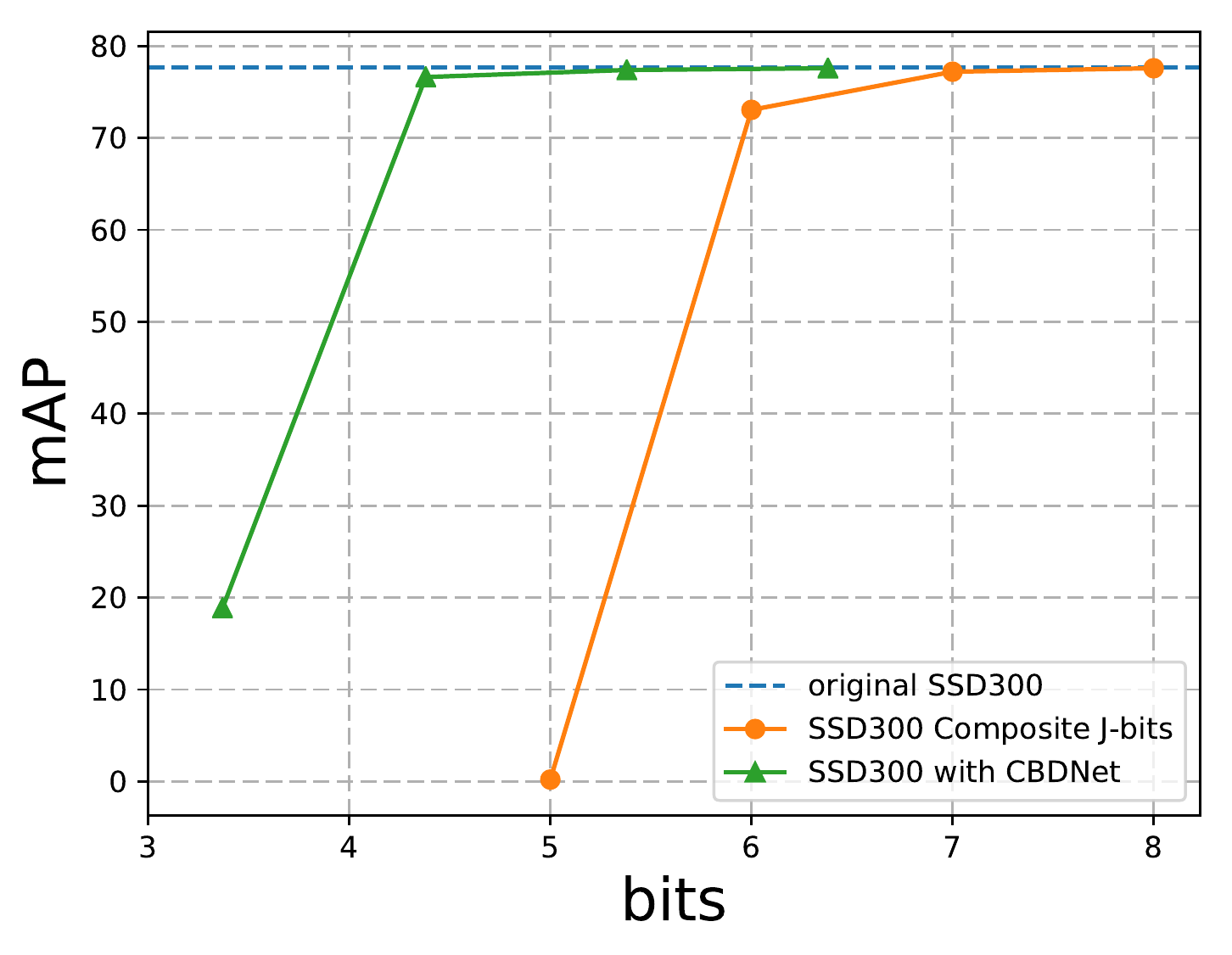}}%
\subfloat[]{\label{fig:segnet}%
\includegraphics[width=.5\linewidth]{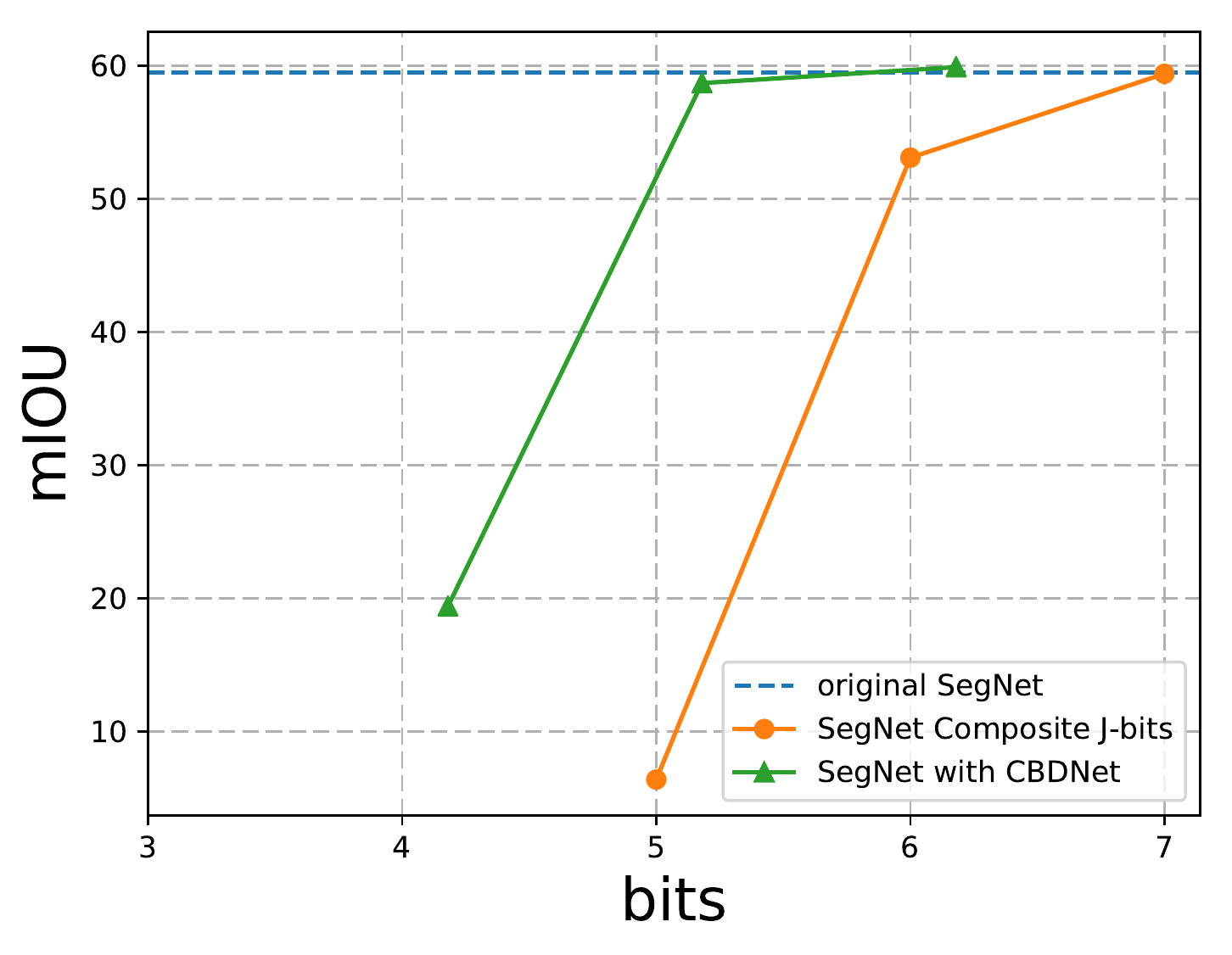}}%
\caption{Accuracy comparison on (a) SSD300 and (b) SegNet.}
\end{minipage}
\end{figure*}

\subsubsection{Model 3:}
\textbf{DenseNet-121} \cite{densenet} consists of 121 layers. The top-1/top-5 accuracy of the FP32 model is $74.41/92.14$  in our single center crop evaluation.
Figure \ref{fig:des} shows the accuracy and the bit-rate of CBDNet at different bottleneck ratios.
The effective bit-rate is 5.72 with $b=0.4$ and $J=7$, while the top-1/top-5 accuracy drops by 1.28\% and 0.85\% respectively.

\subsubsection{Comparison to State-of-the-art}
We further compare the accuracy and bit-rate to the prior art binary weighted decomposition (BWD) \cite{kamiya2017binary}.
Table \ref{tab:comp} lists the comparison results on ResNet-152 and VGG-16.
Note that, BWD keeps the first conv-layer un-decomposed to ensure no significant accuracy drops,
while our CBDNet decomposes all the layers.
It shows that our CBDNet achieves much less accuracy drops even with smaller bit-rate.

\subsection{Detection Networks}
We apply CBDNet to the object detection network SSD300 \cite{Liu2016SSD}.
The evaluation is performed on the VOC0712 dataset. SSD300 is trained with the combined training set from VOC 2007 \texttt{trainval} and VOC 2012 \texttt{trainval} (``07+12''), and tested on the VOC 2007 testset.
We compare the performance between original SSD300 and our CBDNet.
Figure~\ref{fig:ssd} shows the comparison results, in which we also include the composite-only (without binary decomposition) results.

The effective bit-rate of CBDNet is 4.38, with 1.34\% drop of the mean Average Precision (mAP).
That verifies the effectiveness of our CBDNet on object detection networks.

\subsection{Semantic Segmentation Networks}
We also perform an evaluation on the semantic segmentation network SegNet \cite{badrinarayanan2017segnet}.
The experiment is conducted on the Cityscapes dataset \cite{Cordts2016Cityscapes} of the 11 class version, for fair comparison to results by \cite{kamiya2017binary}.
We use the public available SegNet model, and test on the Cityscapes validation set.
Figure~\ref{fig:segnet} shows the comparison results, in which the composite-only results are also included.

\begin{figure}[t!]
\centering
\small
\includegraphics[width=1\linewidth]{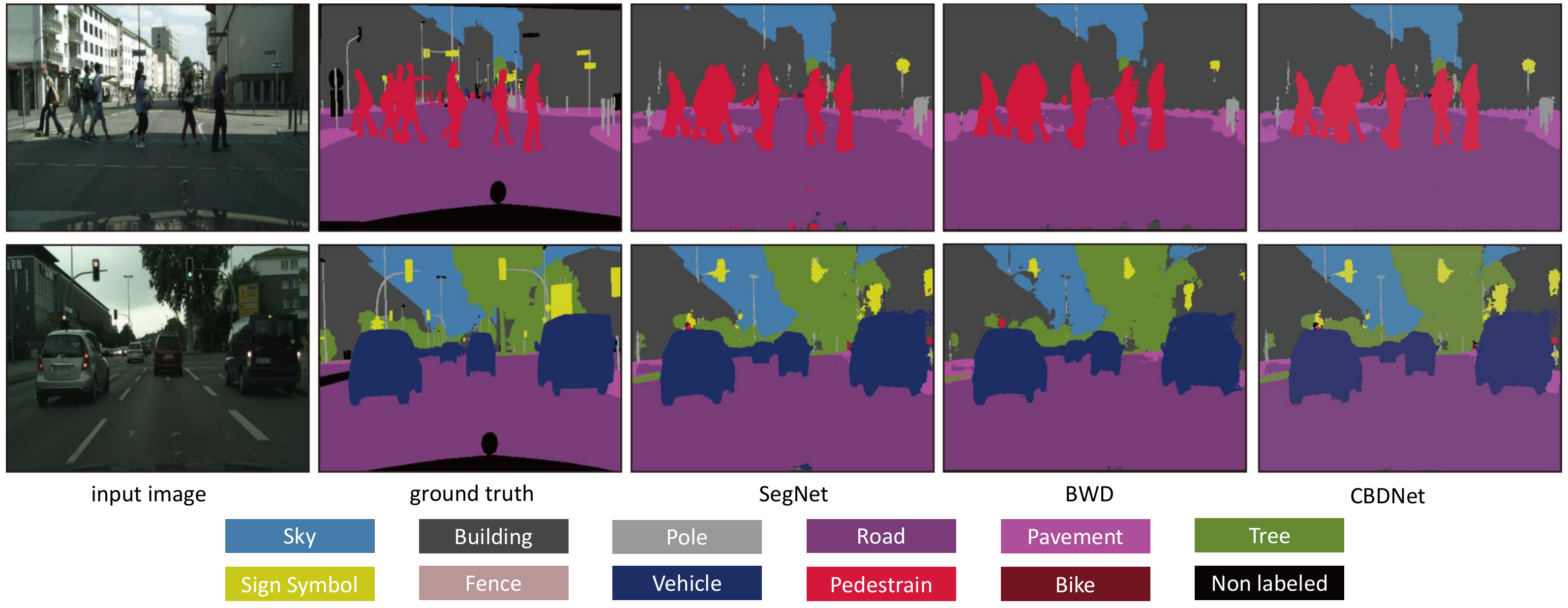}
\caption{Semantic segmentation results on CityScapes. CBDNet is better than BWD on ``sign" and ``pole".
}
\label{fig:segnet-result}
\end{figure}

The effective  bit-rate of CBDNet on SegNet is 5.18, with only 0.8\% drop of the mean of intersection over union (mIOU).
In comparison, under the same setting, the BWD \cite{kamiya2017binary} requires 6 bits but yields more than 2.75\% accuracy loss.
Figure \ref{fig:segnet-result} further illustrates the segmentation results on two tested images, which compares results from
ground truth, SegNet, BWD, and our CBDNet. It is obvious that our CBDNet gives better results than BWD.
That verifies the effectiveness of our CBDNet on semantic segmentation networks.

\subsection{Inference Speed with CBDNet}
We make the inference speed comparison between CBDNet and the FP32 version on Intel Core i7-6700 CPU with 32GB RAM.
We optimize one conv-layer of VGG-16 using SSE4.2 instructions (128-bit SIMD) for both the FP32 version and our CBDNet version.
For CBDNet, we quantize the input into 8-bits using the TensorFlow quantizer \cite{googlewhitepaper}, and then perform bitwise operations.
We project results in this layer to the whole network based on operation numbers per-layer, and show that our CBDNet is about 4.02$\times$ faster than the FP32 counterpart.
In comparison, BWD only gives 2.07$\times$ speedup over the FP32 version with SSE4.2 optimization, under 6-bits data/weight quantization.
This clearly demonstrates our advantage over BWD \cite{kamiya2017binary}.
We believe specific designed hardware with dedicated bitwise accelerators could realize ultra efficient inference for our CBDNet.


\section{Conclusion}
This paper proposes composite binary decomposition networks (CBDNet),
which can directly transfer pre-trained full-precision CNN models into multi-bits binary network models in a training-free way,
while remain model accuracy and computing efficiency.
The method contains two steps, composite real-valued tensors into a limited number of binary tensors, and decomposing certain conditioned binary tensors into two low-rank tensors for parameter and computing efficiency.
Experiments demonstrate the effectiveness of the proposed method on various classification networks like ResNet, DenseNet, VGGNet,
as well as object detection network SSD300, and semantic segmentation network SegNet.

\noindent\textbf{Acknowledgement:}
Y. Qiaoben and J. Zhu were supported by the National Key Research and Development  Program  of  China  (2017YFA0700904), NSFC projects (61620106010, 61621136008, 61332007) and the  MIIT  Grant  of  Int.  Man.  Comp.  Stan (2016ZXFB00001). Yu-Gang Jiang was supported partly by NSFC projects (61622204, U1509206). 

{
    \bibliographystyle{aaai}
    \bibliography{egbib}

\begin{thebibliography}{}

\bibitem[\protect\citeauthoryear{Badrinarayanan \bgroup et al\mbox.\egroup
  }{2017}]{badrinarayanan2017segnet}
Badrinarayanan, V.; Kendall, A.; Cipolla, R.; et~al.
\newblock 2017.
\newblock {SegNet}: A deep convolutional encoder-decoder architecture for image
  segmentation.
\newblock {\em IEEE Trans PAMI} 39.

\bibitem[\protect\citeauthoryear{Chen \bgroup et al\mbox.\egroup
  }{2015}]{hashnet}
Chen, W.; Wilson, J.; S, T.; et~al.
\newblock 2015.
\newblock Compressing neural networks with the hashing trick.
\newblock In {\em ICML}.

\bibitem[\protect\citeauthoryear{Cordts \bgroup et al\mbox.\egroup
  }{2016}]{Cordts2016Cityscapes}
Cordts, M.; Omran, M.; Ramos, S.; et~al.
\newblock 2016.
\newblock The cityscapes dataset for semantic urban scene understanding.
\newblock In {\em CVPR}.

\bibitem[\protect\citeauthoryear{Courbariaux \bgroup et al\mbox.\egroup
  }{2015}]{courbariaux2015binaryconnect}
Courbariaux, M.; Bengio, Y.; David, J.-P.; et~al.
\newblock 2015.
\newblock {Binaryconnect}: Training deep neural networks with binary weights
  during propagations.
\newblock In {\em NIPS}.

\bibitem[\protect\citeauthoryear{Courbariaux \bgroup et al\mbox.\egroup
  }{2016}]{binarynet}
Courbariaux, M.; Bengio, Y.; David, J.-P.; et~al.
\newblock 2016.
\newblock Binarynet: Training deep neural networks with weights and activations
  constrained to+ 1 or-1.
\newblock In {\em ICLR}.

\bibitem[\protect\citeauthoryear{Denton \bgroup et al\mbox.\egroup
  }{2014}]{Denton2014Exploiting}
Denton, E.; Zaremba; Lecun, Y.; et~al.
\newblock 2014.
\newblock Exploiting linear structure within convolutional networks for
  efficient evaluation.
\newblock In {\em NIPS}.

\bibitem[\protect\citeauthoryear{Dong \bgroup et al\mbox.\egroup
  }{2017}]{dong2017sq}
Dong, Y.; Ni, R.; Li, J.; et~al.
\newblock 2017.
\newblock Learning accurate low-bit deep neural networks with stochastic
  quantization.
\newblock In {\em BMVC}.

\bibitem[\protect\citeauthoryear{Girshick \bgroup et al\mbox.\egroup
  }{2014}]{girshick2014rich}
Girshick, R.; Donahue, J.; Darrell, T.; et~al.
\newblock 2014.
\newblock Rich feature hierarchies for accurate object detection and semantic
  segmentation.
\newblock In {\em CVPR}.

\bibitem[\protect\citeauthoryear{Guo \bgroup et al\mbox.\egroup
  }{2018}]{guo2018nd}
Guo, J.; Li, Y.; Lin, W.; et~al.
\newblock 2018.
\newblock Network decoupling: From regular to depthwise separable convolutions.
\newblock In {\em BMVC}.

\bibitem[\protect\citeauthoryear{Han \bgroup et al\mbox.\egroup
  }{2015}]{han-learning}
Han, S.; Pool, J.; Tran, J.; Dally, W.; et~al.
\newblock 2015.
\newblock Learning both weights and connections for efficient neural network.
\newblock In {\em NIPS}.

\bibitem[\protect\citeauthoryear{Han \bgroup et al\mbox.\egroup
  }{2016}]{deep-compression}
Han, S.; Mao, H.; Dally, W.~J.; et~al.
\newblock 2016.
\newblock Deep compression: Compressing deep neural network with pruning,
  trained quantization and huffman coding.
\newblock In {\em ICLR}.

\bibitem[\protect\citeauthoryear{He \bgroup et al\mbox.\egroup
  }{2016}]{he2015deep}
He, K.; Zhang, X.; Ren, S.; and Sun, J.
\newblock 2016.
\newblock Deep residual learning for image recognition.
\newblock In {\em CVPR}.

\bibitem[\protect\citeauthoryear{Hinton \bgroup et al\mbox.\egroup
  }{2015}]{hinton2015distilling}
Hinton, G.; Vinyals, O.; Dean, J.; et~al.
\newblock 2015.
\newblock Distilling the knowledge in a neural network.
\newblock {\em arXiv preprint arXiv:1503.02531}.

\bibitem[\protect\citeauthoryear{Hou \bgroup et al\mbox.\egroup
  }{2017}]{hou2016loss}
Hou, L.; Yao, Q.; Kwok, J.~T.; et~al.
\newblock 2017.
\newblock Loss-aware binarization of deep networks.
\newblock In {\em ICLR}.

\bibitem[\protect\citeauthoryear{Howard \bgroup et al\mbox.\egroup
  }{2017}]{howard2017mobilenets}
Howard, A.~G.; Zhu, M.; Chen, B.; et~al.
\newblock 2017.
\newblock Mobilenets: Efficient convolutional neural networks for mobile vision
  applications.
\newblock {\em arXiv preprint arXiv:1704.04861}.

\bibitem[\protect\citeauthoryear{Huang \bgroup et al\mbox.\egroup
  }{2017}]{densenet}
Huang, G.; Liu, Z.; Weinberger, K.~Q.; et~al.
\newblock 2017.
\newblock Densely connected convolutional networks.
\newblock In {\em CVPR}.

\bibitem[\protect\citeauthoryear{Jaderberg \bgroup et al\mbox.\egroup
  }{2014}]{Jaderberg2014Speeding}
Jaderberg, M.; Vedaldi, A.; Zisserman, A.; et~al.
\newblock 2014.
\newblock Speeding up convolutional neural networks with low rank expansions.
\newblock In {\em BMVC}.

\bibitem[\protect\citeauthoryear{Kamiya \bgroup et al\mbox.\egroup
  }{2017}]{kamiya2017binary}
Kamiya, R.; Yamashita, T.; Ambai, M.; et~al.
\newblock 2017.
\newblock Binary-decomposed dcnn for accelerating computation and compressing
  model without retraining.
\newblock In {\em ICCV Workshop}.

\bibitem[\protect\citeauthoryear{Krishnamoort}{2018}]{googlewhitepaper}
Krishnamoort, R.
\newblock 2018.
\newblock Quantizing deep convolutional networks for efficient inference: A
  whitepaper.
\newblock {\em arXiv preprint arXiv:1805.07941}.

\bibitem[\protect\citeauthoryear{Krizhevsky \bgroup et al\mbox.\egroup
  }{2012}]{krizhevsky2012imagenet}
Krizhevsky, A.; Sutskever, I.; Hinton, G.~E.; et~al.
\newblock 2012.
\newblock Imagenet classification with deep convolutional neural networks.
\newblock In {\em NIPS}.

\bibitem[\protect\citeauthoryear{Li \bgroup et al\mbox.\egroup
  }{2017}]{Li2016Pruning}
Li, H.; Kadav, A.; I, D.; et~al.
\newblock 2017.
\newblock Pruning filters for efficient convnets.
\newblock In {\em ICLR}.

\bibitem[\protect\citeauthoryear{Lin \bgroup et al\mbox.\egroup
  }{2017}]{lin2017abc}
Lin, X.; Zhao, C.; Pan, W.; et~al.
\newblock 2017.
\newblock Towards accurate binary convolutional neural network.
\newblock In {\em NIPS}.

\bibitem[\protect\citeauthoryear{Liu \bgroup et al\mbox.\egroup
  }{2016}]{Liu2016SSD}
Liu, W.; Anguelov, D.; Erhan, D.; et~al.
\newblock 2016.
\newblock {SSD}: Single shot multibox detector.
\newblock In {\em ECCV}.

\bibitem[\protect\citeauthoryear{Liu \bgroup et al\mbox.\egroup
  }{2017a}]{liu2017pnas}
Liu, C.; Zoph, B.; Neumann, M.; et~al.
\newblock 2017a.
\newblock Progressive neural architecture search.
\newblock {\em arXiv preprint arXiv:1712.00559}.

\bibitem[\protect\citeauthoryear{Liu \bgroup et al\mbox.\egroup
  }{2017b}]{liu2017learning}
Liu, Z.; Li, J.; Shen, Z.; et~al.
\newblock 2017b.
\newblock Learning efficient convolutional networks through network slimming.
\newblock In {\em ICCV}.

\bibitem[\protect\citeauthoryear{Long \bgroup et al\mbox.\egroup
  }{2015}]{long2015fully}
Long, J.; Shelhamer, E.; Darrell, T.; et~al.
\newblock 2015.
\newblock Fully convolutional networks for semantic segmentation.
\newblock In {\em CVPR}.

\bibitem[\protect\citeauthoryear{Miettinen}{2010}]{miettinen2010sparse}
Miettinen, P.
\newblock 2010.
\newblock Sparse boolean matrix factorizations.
\newblock In {\em ICDM}.

\bibitem[\protect\citeauthoryear{Migacz}{2017}]{tensorrt}
Migacz, S.
\newblock 2017.
\newblock 8-bit inference with {T}ensor{RT}.
\newblock Technical report, NVidia.

\bibitem[\protect\citeauthoryear{Rastegari \bgroup et al\mbox.\egroup
  }{2016}]{rastegari2016xnor}
Rastegari, M.; Ordonez, V.; Redmon, J.; and Farhadi, A.
\newblock 2016.
\newblock {XNOR-Net}: Imagenet classification using binary convolutional neural
  networks.
\newblock In {\em ECCV}.

\bibitem[\protect\citeauthoryear{Ren \bgroup et al\mbox.\egroup
  }{2015}]{ren2015faster}
Ren, S.; He, K.; Girshick, R.; and Sun, J.
\newblock 2015.
\newblock Faster {R-CNN}: Towards real-time object detection with region
  proposal networks.
\newblock In {\em NIPS}.

\bibitem[\protect\citeauthoryear{Russakovsky \bgroup et al\mbox.\egroup
  }{2015}]{ILSVRC15}
Russakovsky, O.; Deng, J.; Su, H.; Krause, J.; et~al.
\newblock 2015.
\newblock {ImageNet Large Scale Visual Recognition Challenge}.
\newblock {\em IJCV} 115(3).

\bibitem[\protect\citeauthoryear{Shen \bgroup et al\mbox.\egroup
  }{2017}]{shen2017dsod}
Shen, Z.; Liu, Z.; Li, J.; et~al.
\newblock 2017.
\newblock {DSOD}: Learning deeply supervised object detectors from scratch.
\newblock In {\em ICCV}.

\bibitem[\protect\citeauthoryear{Simonyan and
  Zisserman}{2015}]{simonyan2014very}
Simonyan, K., and Zisserman, A.
\newblock 2015.
\newblock Very deep convolutional networks for large-scale image recognition.
\newblock In {\em ICLR}.

\bibitem[\protect\citeauthoryear{Szegedy \bgroup et al\mbox.\egroup
  }{2015}]{szegedy2015going}
Szegedy, C.; Liu, W.; Jia, Y.; et~al.
\newblock 2015.
\newblock Going deeper with convolutions.
\newblock In {\em CVPR}.

\bibitem[\protect\citeauthoryear{Wen \bgroup et al\mbox.\egroup
  }{2016}]{Wen2016Learning}
Wen, W.; Wu, C.; Wang, Y.; et~al.
\newblock 2016.
\newblock Learning structured sparsity in deep neural networks.
\newblock In {\em NIPS}.

\bibitem[\protect\citeauthoryear{Zhang \bgroup et al\mbox.\egroup
  }{2007}]{zhang2007binary}
Zhang, Z.; Li, T.; Ding, C.; and Zhang, X.
\newblock 2007.
\newblock Binary matrix factorization with applications.
\newblock In {\em ICDM}.

\bibitem[\protect\citeauthoryear{Zhang \bgroup et al\mbox.\egroup
  }{2016}]{Zhang2016Accelerating}
Zhang, X.; Zou, J.; Sun, J.; et~al.
\newblock 2016.
\newblock Accelerating very deep convolutional networks for classification and
  detection.
\newblock {\em IEEE Trans PAMI} 38(10).

\bibitem[\protect\citeauthoryear{Zhang \bgroup et al\mbox.\egroup
  }{2017}]{zhang2017igc}
Zhang, T.; Qi, G.; Wang, J.; et~al.
\newblock 2017.
\newblock Interleaved group convolutions.
\newblock In {\em ICCV}.

\bibitem[\protect\citeauthoryear{Zhang \bgroup et al\mbox.\egroup
  }{2018}]{zhang2017shufflenet}
Zhang, X.; Zhou, X.; Sun, J.; et~al.
\newblock 2018.
\newblock Shufflenet: An extremely efficient convolutional neural network for
  mobile devices.
\newblock In {\em CVPR}.

\bibitem[\protect\citeauthoryear{Zhou \bgroup et al\mbox.\egroup
  }{2016}]{zhou2016dorefa}
Zhou, S.; Wu, Y.; Ni, Z.; et~al.
\newblock 2016.
\newblock {DoReFa-Net}: Training low bitwidth convolutional neural networks
  with low bitwidth gradients.
\newblock {\em arXiv preprint arXiv:1606.06160}.

\bibitem[\protect\citeauthoryear{Zoph and Le}{2016}]{zoph2016neural}
Zoph, B., and Le, Q.~V.
\newblock 2016.
\newblock Neural architecture search with reinforcement learning.
\newblock In {\em ICLR}.

\end{thebibliography}
}
\end{document}